\documentclass[acmsmall,screen]{acmart}

\usepackage[linesnumbered,ruled,vlined]{algorithm2e}
\usepackage{amsmath}
\usepackage{subfig}
\usepackage{xcolor}
\usepackage{bm}

\usepackage{graphicx}
\usepackage{multirow}
\usepackage{mdframed}
\usepackage{bm}
\usepackage{makecell}
\usepackage{mathtools}
\usepackage{wasysym}
\usepackage{arydshln}
\usepackage{caption}
\usepackage{color}
\usepackage{CJKutf8}

\usepackage{paralist}

\usepackage{microtype}

\definecolor{ipink}{RGB}{245,39,209}
\definecolor{iblue}{RGB}{139,106,255}

\newcommand{\paratitle}[1]{\vspace{1.5ex}\noindent \textbf{#1}}
\makeatletter
\newcommand*{\da@rightarrow}{\mathchar"0\hexnumber@\symAMSa 4B }
\newcommand*{\da@leftarrow}{\mathchar"0\hexnumber@\symAMSa 4C }
\newcommand*{\xdashrightarrow}[2][]{%
  \mathrel{%
    \mathpalette{\da@xarrow{#1}{#2}{}\da@rightarrow{\,}{}}{}%
  }%
}
\newcommand{\xdashleftarrow}[2][]{%
  \mathrel{%
    \mathpalette{\da@xarrow{#1}{#2}\da@leftarrow{}{}{\,}}{}%
  }%
}
\newcommand*{\da@xarrow}[7]{%

  \sbox0{$\ifx#7\scriptstyle\scriptscriptstyle\else\scriptstyle\fi#5#1#6\m@th$}%
  \sbox2{$\ifx#7\scriptstyle\scriptscriptstyle\else\scriptstyle\fi#5#2#6\m@th$}%
  \sbox4{$#7\dabar@\m@th$}%
  \dimen@=\wd0 %
  \ifdim\wd2 >\dimen@
    \dimen@=\wd2 %
  \fi
  \count@=2 %
  \def\da@bars{\dabar@\dabar@}%
  \@whiledim\count@\wd4<\dimen@\do{%
    \advance\count@\@ne
    \expandafter\def\expandafter\da@bars\expandafter{%
      \da@bars
      \dabar@
    }%
  }%
  \mathrel{#3}%
  \mathrel{%
    \mathop{\da@bars}\limits
    \ifx\\#1\\%
    \else
      _{\copy0}%
    \fi
    \ifx\\#2\\%
    \else
      ^{\copy2}%
    \fi
  }%
  \mathrel{#4}%
}
\makeatother

\SetCommentSty{mycommfont}

\SetKwInput{KwInput}{Input}                
\SetKwInput{KwOutput}{Output}              

\AtBeginDocument{%
  \providecommand\BibTeX{{%
    \normalfont B\kern-0.5em{\scshape i\kern-0.25em b}\kern-0.8em\TeX}}}

\setcopyright{acmcopyright}
\copyrightyear{2023}
\acmYear{2023}
\acmDOI{XXXXXXX.XXXXXXX}

\acmVolume{37}
\acmNumber{4}
\acmArticle{1}
\acmMonth{8}

\begin{document}

\title{Revisiting Conversation Discourse for Dialogue Disentanglement}


\author{Bobo Li}
\email{boboli@whu.edu.cn}
\affiliation{%
  \institution{Wuhan University}
  \country{China}
}

\author{Hao Fei}
\affiliation{%
  \institution{National University of Singapore}
  \city{Singapore}
  \country{Singapore}}
\email{haofei37@nus.edu.sg}

\author{Fei Li}
\email{lifei_csnlp@whu.edu.cn}
\affiliation{%
  \institution{Wuhan University}
  \country{China}
}

\author{Shengqiong Wu}
\affiliation{%
  \institution{National University of Singapore}
  \country{Singapore}
  }

\author{Lizi Liao}
\affiliation{%
  \institution{Singapore Management University}
  \city{Singapore}
  \country{Singapore}
}

\author{Yinwei Wei}
\affiliation{%
  \institution{National University of Singapore}
  \country{Singapore}
  }

\author{Tat-seng Chua}
\affiliation{%
  \institution{National University of Singapore}
  \country{Singapore}
  }

\author{Donghong Ji}
\email{dhji@whu.edu.cn}
\affiliation{%
  \institution{Wuhan University}
  \country{China}
}

\renewcommand{\shortauthors}{Li, et al.}

\begin{abstract}
Dialogue disentanglement aims to detach the chronologically ordered utterances into several independent sessions.
Conversation utterances are essentially organized and described by the underlying discourse, and thus dialogue disentanglement requires the full understanding and harnessing of the intrinsic discourse attribute.
In this paper, we propose enhancing dialogue disentanglement by taking full advantage of the dialogue discourse characteristics.
First of all, \textbf{in feature encoding stage}, we construct the heterogeneous graph representations to model the various dialogue-specific discourse structural features, including the static speaker-role structures (i.e., speaker-utterance and speaker-mentioning structure) and the dynamic contextual structures (i.e., the utterance-distance and partial-replying structure).
We then develop a structure-aware framework to integrate the rich structural features for better modeling the conversational semantic context.
Second, \textbf{in model learning stage}, we perform optimization with a hierarchical ranking loss mechanism, which groups dialogue utterances into different discourse levels and carries training covering pair-wise and session-wise levels hierarchically.
Third, \textbf{in inference stage}, we devise an easy-first decoding algorithm, which performs utterance pairing under the easy-to-hard manner with a global context, breaking the constraint of traditional sequential decoding order.
On two benchmark datasets, our overall system achieves new state-of-the-art performances on all evaluations.
In-depth analyses further demonstrate the efficacy of each proposed idea and also reveal how our methods help advance the task.
Our work has great potential to facilitate broader multi-party multi-thread dialogue applications.
\end{abstract}



\maketitle

\section{Introduction}

\begin{figure}[!t]
  \centering
  \includegraphics[width=0.95\columnwidth]{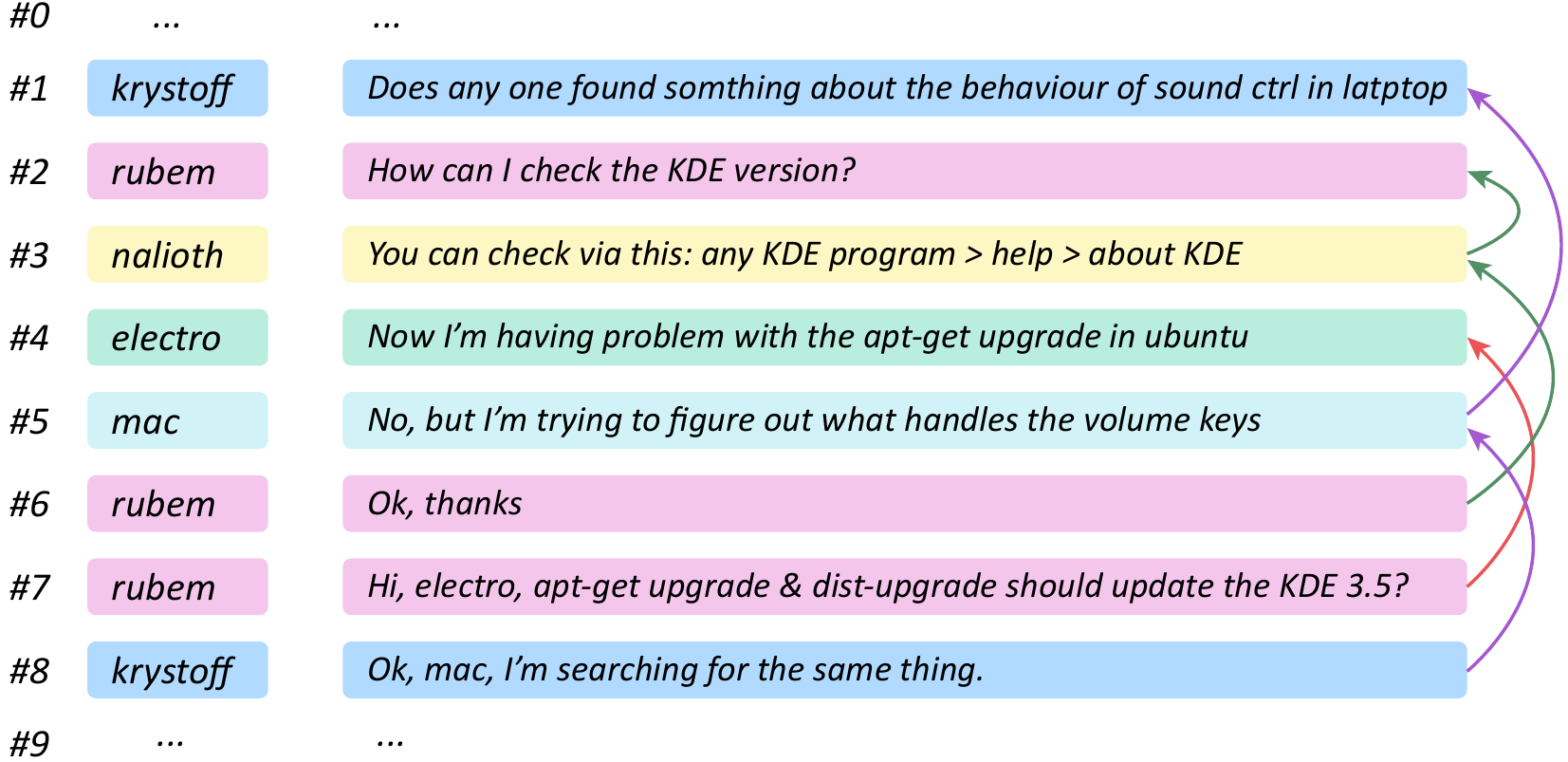}
  \caption{
A conversation snippet with the replying structure.
Utterances from one speaker are marked with the same color.
Utterances linked by same-color arrows have shared sessions.
  }
  \label{fig:intro}
\end{figure}

Multi-turn multi-party conversations are often characterized by intertwined utterances with many speakers, and multiple coexisted topic threads \cite{traum-etal-2004-evaluation}.
This adds challenges to the dialogue understanding and responding.
Dialogue disentanglement has thus been proposed with the aim of decomposing entangled utterances from different threads or sessions \cite{kummerfeld-etal-2019-large}, i.e., finding the \emph{reply-to} relations between the chronologically-listed utterances (cf. Figure \ref{fig:intro}).
Earlier research builds handcrafted discrete features with machine learning models to cluster utterances into different sessions or predict whether there is a replying relation between utterances \cite{elsner-charniak-2008-talking,elsner-charniak-2010-disentangling,icwsm-AumayrCH11}.
Currently, the rapid development of deep neural models has greatly advanced the dialogue disentanglement task \cite{mehri-carenini-2017-chat,liuSGLWZ20-ijcai,jiang-etal-2018-identifying,chengyuhuang-emnlp}, especially by making use of the pre-trained language models (PLM) \cite{aaai-ZhuNWNX20,tiandali-arxiv}, e.g., BERT \cite{devlin-etal-2019-bert}, DiaBERT \cite{tiandali-arxiv}.
As extensively revealed, the essence of the task lies in the understanding of the underlying conversational discourse \cite{zhu-etal-2021-findings,aclMa0Z22}, and thus it is key to model the discourse structure of the dialogue.
Despite the progress of dialogue disentanglement achieved by prior efforts, existing explorations, unfortunately, still need to harness the conversational discourse nature fully.

At \emph{feature modeling} perspective, current research fails to sufficiently make use of the conversational discourse structural information.
In the dialogue context, there are various types of discourse structures beneficial to the dialogue disentanglement task.
The first type is the \textbf{speaker-utterance discourse}.
In the multi-party dialog, due to the participant persona consistency, different utterances by the same speaker may exhibit identical styles.
Thus, capturing the speaker-utterance correlations will facilitate the responding recognition.
Second, the \textbf{speaker-mentioning discourse}.
The speaker coreferences indicate the interactions between different utterances and speakers, which are the value clues to the detection of replying relation.
Third, the \textbf{utterance-distance discourse}.
A conversation usually involves a great number of turns,
in which the valid contexts that offer critical features for the replying reasoning actually center around the current utterance (i.e., near neighbors), and the farther away, the lower the efficacy of the features.
The Ubuntu IRC data \cite{kummerfeld-etal-2019-large} statistics shows that the average turn of conversation is 10.17, while 80\%/90\%/95\% replying relations are scattered within 6/13/21 utterances forward and backward, respectively.
Fourth, the \textbf{partial-replying discourse}.
In fact, the incorporation of the previously-discovered partial replying structure is beneficial to the detection of the following replying relation.
For example as shown in Figure \ref{fig:intro}, directly determining the reply-to relation \#6$^{\curvearrowright}$\#3 can be tricky; while knowing the replying relation of \#3$^{\curvearrowright}$\#2 as prior, the detection of \#6$^{\curvearrowright}$\#3 can be greatly eased.

At \emph{system optimization} perspective, existing works disregard the dialogue discourse characteristic for model training and decoding.
On the one hand, utterances are governed under both the local replying thread and global session discourses, and thus, the dialogue disentanglement task measures both the session-level and pair-wise detection.
Unfortunately, most current models only train with the pair-wise cross-entropy loss without considering the higher-level optimization (e.g., thread, session) \cite{liuSGLWZ20-ijcai,jiang-etal-2018-identifying,aclMa0Z22}.
On the other hand, existing methods take a front-to-end reading order to decode the replying relation for an utterance with merely the precedent context.
Yet this decoding approach can be less effective and provide less informative results as the semantics of a conversation are organized in a hierarchical structure rather than a linear chronological order.
Intrinsically, as humans, we always first recognize those utterances with simple replying relations, then try the harder cases gradually with more clues.
Taking Figure \ref{fig:intro} as an example with \#5 as the current utterance, it can be easier to first determine the replying dependency of \#8$^{\curvearrowright}$\#5 with the cue word `\emph{mac}' in the following context.
Based on the established replying structure \#8$^{\curvearrowright}$\#5, the hard one of \#5$^{\curvearrowright}$\#1 can be further detected without much effort.

\begin{figure}[!t]
  \centering
  \includegraphics[width=0.99\columnwidth]{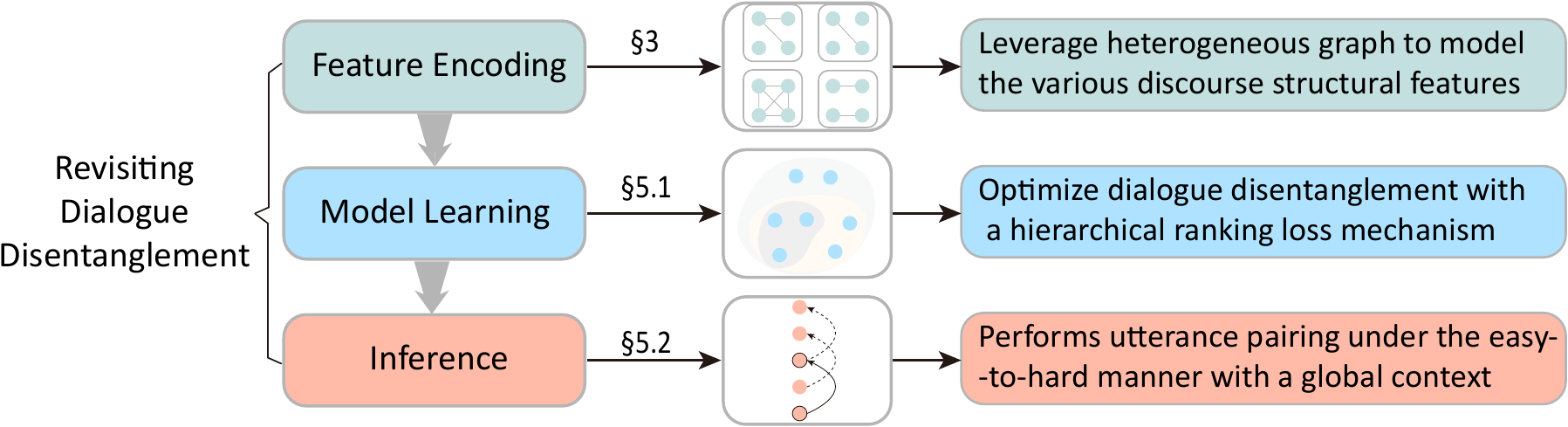}
  \caption{
  Summary of our main proposals for enhancing the dialogue disentanglement in different stages.
  }
  \label{fig:summary}
\end{figure}

In this work, we rethink the conversation discourse to dialogue disentanglement, and propose to enhance the task by giving full consideration to the above observations.
First of all, to take advantage of the rich conversational discourse features ($\S$\ref{Construction of Dialogue Discourse Structures}), we construct the four types of graphs to represent the aforementioned discourse structures.
We consider the two static speaker-role graphs, including the speaker-utterance and speaker-mentioning structures.
Also, we build the two dynamic contextual structures, including a Gaussian-based utterance-distance structure and a dynamically updated partial-replying structure.
We further develop a structure-aware framework ($\S$\ref{Structure-aware Framework}) for dialogue disentanglement.
As shown in Figure \ref{fig:model}, we encode and integrate the various heterogeneous graphs with edge-aware graph convolutional networks (EGCN), where the resulting rich structural features aid the modeling of the intrinsic conversational contexts.
Figure~\ref{fig:summary} illustrates our enhancement for dialogue disentanglement in different stages: \textit{feature encoding}, \textit{model learning}, and \textit{inference}.

Further, we propose to optimize the learning and inference of the above framework, following the hierarchical nature of conversation discourse.
First, for model training, we devise a hierarchical ranking loss mechanism (cf $\S$\ref{sec:hi-ranking}).
We group the candidate parents of the current utterance into different discourse levels within a dialogue, based on which we define the learning losses under three hierarchical levels, covering both the pair-wise and session-wise optimizations.
Then, during inference, we introduce an easy-first relation decoding algorithm (cf $\S$\ref{sec:Easy-First Graph-based Decoding}).
By consulting both the precedent and subsequent context, the utterance-parent pairing procedure is taken place in an easy-to-hard manner without following the sequential order.
Specifically, we maintain a global utterance-pair scoring matrix, and at each decoding step, the utterance pair with the highest score will be selected.
Also, the established replying pair is incrementally added into the partial-replying structure to update features for further facilitating the follow-up inference.

We conduct experiments on two benchmark datasets, including the Ubuntu IRC \cite{kummerfeld-etal-2019-large} and Movie Dialogue \cite{liuSGLWZ20-ijcai}.
The results show that our overall system outperforms the current state-of-the-art (SoTA) baselines with significant margins on all datasets and metrics.
Model ablation studies prove the necessity of integrating the various dialogue discourse structure information, the hierarchical ranking loss mechanism, and also the easy-first decoding algorithm for dialogue disentanglement.
Additionally, our experiments highlight the effectiveness of our discourse structure-aware method, particularly in scenarios with longer utterance distance and multiple participating speakers.
Furthermore, our in-depth analysis of the Hierarchical Ranking Loss mechanism reveals its crucial role in rectifying prediction errors and improving dialogue disentanglement performance.
The insights derived from our experimental analysis also underscore the power of the easy-first decoding strategy in facilitating confident decision-making, thereby demonstrating its significant advantage in the field of dialogue disentanglement.

Moreover, we conducted an experiment comparing our model to GPT-3.5 for the task of dialogue disentanglement.
The results of our experiment demonstrate a clear superiority of our model over GPT-3.5, indicating a significant advancement in the dialogue disentanglement task.
These outcomes suggest that larger language models, despite their comprehensive capabilities, exhibit discernible shortcomings in the dialogue disentanglement task, particularly with respect to effectively capturing dialogue discourse structures.
Based on these insights, we validate the effectiveness of our model and emphasize the necessity for specialized optimization of models tailored for dialogue discourse.

All in all, this paper revisits the discourse attribute of conversation for better dialogue disentanglement.
To our knowledge, this is by far the first work taking full consideration of dialogue discourse, from feature modeling to model optimization.
To aid understanding, we summarize our key contributions as follows.
\begin{itemize}
    \item We construct various dialogue discourse structural features to enrich the dialogue contexts.
    \item We propose a hierarchical ranking loss method to cover both pair-wise and session-wise task learning.
    \item We present an easy-first decoding algorithm to enable a highly effective inference of replying relations. 
\end{itemize}
Our work can be instructive for a wider range of multi-party multi-thread dialogue applications without much effort.
To facilitate the follow-up research, we will release all our codes and metadata upon acceptance.


\section{Relate Work}
\subsection{Dialogue Disentanglement}
Dialogue disentanglement, also known as conversation disentanglement, conversation management, or thread detection \cite{elsner-charniak-2008-talking, traum-etal-2004-evaluation,sigir-ShenYSC06,elsner-charniak-2010-disentangling}, has long been an important research topic in the field of dialogue understanding.
Dialogue disentanglement is also the prerequisite for a wide range of downstream applications, such as
response generation \cite{liangMZCX021-aaai,gu-etal-2022-hetermpc},
dialogue state tracking \cite{ZhangLHZC19,ouyang-etal-2020-dialogue},
dialog-based machine reading comprehension \cite{ouyang-etal-2021-dialogue} and
dialogue information extraction \cite{ghodag-EMNLP-2019,haofei-global}.
The task has been extensively modeled as a pair-wise relation classification, i.e., determining the rely-to relation between utterance pairs.
Early works rely heavily on discrete handcrafted features with machine learning models \cite{elsner-charniak-2008-talking,elsner-charniak-2010-disentangling,icwsm-AumayrCH11}, modeling the utterances into the pair-wise relation prediction.
Later research frequently utilizes the deep learning models to train classifiers \cite{kummerfeld-etal-2019-large,tan-etal-2019-context,liuSGLWZ20-ijcai,zhu-etal-2021-findings}, especially with the use of contextual features from PLMs \cite{aaai-ZhuNWNX20,tiandali-arxiv}.
For example, ~\citet{mehri-carenini-2017-chat,jiang-etal-2018-identifying} transform the dialogue disentanglement into a link prediction problem with LSTM and CNN models.
\citet{yu-joty-2020-online} leverage the pointer network to achieve online decoding.
\citet{chengyuhuang-emnlp} propose to learn the utterance-level and session-level representation based on contrastive learning.

Representing the key properties of dialogue utterances with effective features is pivotal to dialogue disentanglement.
Prior research has explored various conversational features, e.g., time, speakers, topics, and dialogue contexts \cite{tiandali-arxiv}, at different levels, e.g., inter-utterance features \cite{yu-joty-2020-online} and session-level features \cite{chengyuhuang-emnlp}.
More recently, \citet{aclMa0Z22} emphasize the learning of conversational discourse information for the task.
The dialogue discourse structure depicts the intrinsic semantic layout and organization of the dialogues.
By integrating the speaker-utterance and speaker-mentioning structural features to enhance the discourse understanding, they thus achieve the current SoTA performance.

This work follows the same line and extends the triumph of \citet{aclMa0Z22} on the modeling of discourse structure information,
Yet ours is more advanced than theirs in three major aspects.
Firstly, \citet{aclMa0Z22} consider merely the two types of static speaker-role structures, which may lead to limited performance improvement.
In contrast, we further consider two dynamic context structures.
Specifically, the utterance-distance structure helps instruct the model to allocate correct attention to utterances according to the cross-utterance distances instead of equal treatment as in \citet{aclMa0Z22}.
And the partial replying structure capturing the direct task clues are overlooked by the existing works.
Secondly, we fully consider the hierarchical discourse structure of the utterance nodes, optimizing the system on the session cluster-level performances with a novel hierarchical ranking loss, instead of the sorely pair-wise measuring in \citet{aclMa0Z22}.
Thirdly, we develop an easy-first decoding algorithm to break the constraint of the traditional front-to-end manner with only the precedent context.
We borrow the easy-first strategies from the syntax parsing community \cite{goldberg-elhadad-2010-efficient,kurita-sogaard-2019-multi}, where the most confident transition decision will be selected at current step based on the global-level context, instead of the uni-directional decoding direction.

\subsection{Discourse Structure Modeling}

To gain a deeper understanding of conversations, it is essential to go beyond semantic content and incorporate the discourse structure, which encompasses elements such as speaker roles, replying relations, and dialogue threading. Analyzing a conversation's discourse structure poses a greater challenge compared to examining flattened documents that only exhibit linear structures. Consequently, researchers have endeavored to explore accurate methods of encoding discourse features, which can be broadly classified into three categories: hierarchical structure modeling, speaker-oriented modeling, and graph-based modeling.

Hierarchical structure modeling, which captures the macro-to-micro hierarchy from the overall dialogue down to specific threads, utterances, and words, is the first among these. This structure offers a valuable perspective for understanding dialogue. Consequently, hierarchical modeling has found widespread application in diverse dialogue-related tasks such as dialogue emotion recognition~\cite{ruanham-ica-2022, lihat-coling-2020}, dialogue state tracking~\cite{rensaa-emnlp-2019, ligae-emnlp-2021, zhoudst-cikm-2022, qicmc-ica-2022}, dialogue-based question answering~\cite{liurtb-acl-2019, littl-acl-2020}, and dialogue generation~\cite{serahl-aaai-2017, zengdlv-emnlp-2019, konghah-ica-2021}, to name a few.

Secondly, the role of the speaker is pivotal in dialogue modeling. Sentences uttered by the same speaker often demonstrate consistency. Moreover, the interpretation of an utterance can vary significantly depending on the speaker to whom it is addressed. In light of these complexities, a range of models have been proposed to exploit the role of the speaker in dialogue.
DialogueRNN~\cite{majdaa-aaai-2019} was among the first to model dialogues from the speaker's perspective, thereby better capturing the speaker's emotions and states. SCIJE~\cite{nansdp-coling-2022} utilizes speaker-aware interactions to model different and similar speakers in the context of conversation representations. Dynamic speaker modeling has also found applications in other tasks, such as dialogue act classification~\cite{hestm-emnlp-2021} and speaker classification~\cite{mengtns-aaai-2018, haoads-naacl-2019}.

Third, discourse graph modeling provides a new perspective for dialogue modeling involving speakers, replying relations, co-reference, and multi-threads. This integration into a graph architecture promises a more comprehensive and precise dialogue modeling. DialogueGCN~\cite{ghodag-EMNLP-2019} considers speaker-level interaction and utterance relative distance for emotion recognition. D2G~\cite{haofei-global} jointly considers speaker and context features for dialogue relation extraction. DSGFNet~\cite{feng-etal-2022-dynamic} uses a dynamic scheme for encoding multi-domain state tracking tasks. 

Given the promising applications of graph-based methods in numerous dialogue-related tasks, this study also employs a graph-based model for dialogue discourse modeling. Distinguishing from previous studies that typically consider only one or two kinds of edges for modeling, we incorporate four kinds of edges to model discourse features, offering the most comprehensive approach for discourse feature exploration and a deep understanding of dialogue structures.

\section{Preliminary}

\subsection{Task Formulation}
Given a dialogue $U$=$\{u_1,\cdots, u_n\}$ in chronological order, each utterance $u_i$=<$s_i,o_i$> consists of the speaker $s_i$ and text content $o_i$.
Also we define a thread set $T$=$\{t_1,\cdots,t_p\}$ as a partition of $U$, with $t_p$=$\{t_{p1},\cdots, t_{pk}\}$ representing a thread of dialogue.
Our goal is to find all reply-to relations between utterances $y_k=\{u_i^{\curvearrowright}u_j\}$ ($i\ge j$), via which we disentangle the $U$ into $T$.

\begin{figure}[!t]
  \centering
  \includegraphics[width=0.98\columnwidth]{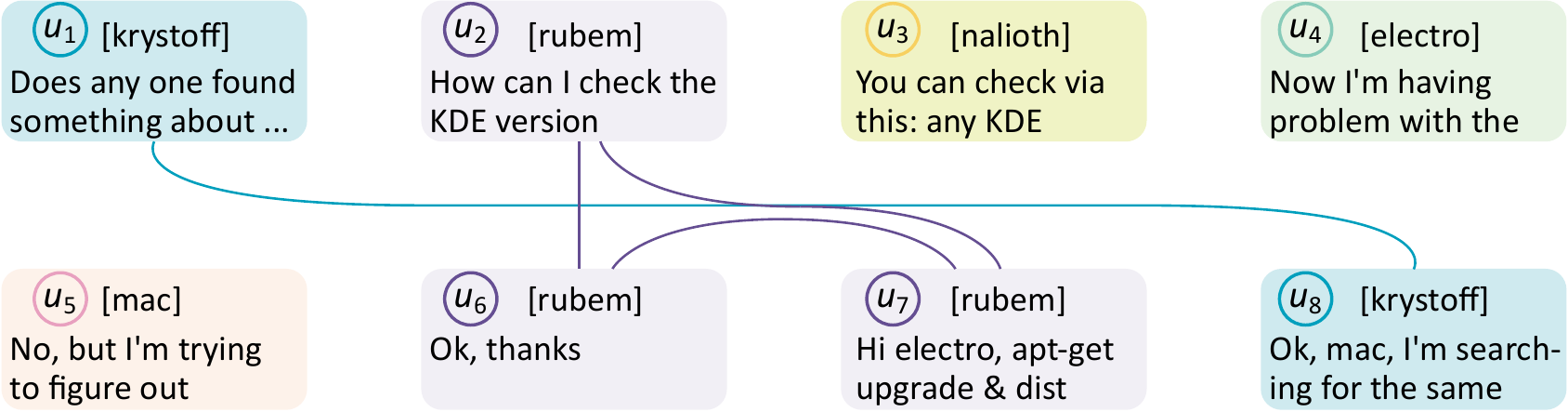}
  \caption{
  Examples of the speaker-utterance structures, where the utterances with the same speaker are connected.
  }
  \label{fig:stat-struct}
\end{figure}


\subsection{Construction of Conversational Discourse Structures}
\label{Construction of Dialogue Discourse Structures}
To facilitate the identification of reply-to relations and to provide a robust representation of dialogue dynamics, we propose constructing four types of dialogue discourse structures.
The first two types, the speaker-utterance structure (denoted as $G^{S}$=<$V, E^{S}$>) and speaker-mentioning structure ($G^{M}$=<$V, E^{M}$>), capture the static speaker roles. On the other hand, the remaining two types, the utterance-distance structure ($G^{D}$=<$V, E^{D}$>) and partial-replying structure ($G^{R}$=<$V, E^{R}$>), describe the dynamic discourse contexts. 
These structures derive four heterogeneous graphs, with the same set $V$ of utterance node $u_i$, but different definitions of edges $E^{S/M/D/R}$.
To strengthen the message passing, we also make four graphs bidirectional, and add the self-loop link for all nodes.

\paratitle{Speaker-utterance Structure.}
Throughout a multi-party dialogue, each participant speaks multiple utterances, which are interspersed with utterances from other speakers.
Generally, due to the persona, different utterances by the same speaker may exhibit a consistent style.
The interaction between these utterances can help capture their consistency and contribute to the overall understanding of the dialogue.
To represent this interaction, we create speaker-utterance edges, linking all utterances generated by the same speakers. 
For example, as shown in Figure \ref{fig:stat-struct}, we connect $u_1$ and $u_8$, both uttered by \textbf{krystoff}. 
Similarly, for utterances $u_2, u_6,$ and $u_8$ from the same speaker \textbf{rubem}, we add links between each pair. 
Technically, we denote an edge $e^S_{i,j}=1$ in ${E}^S$ if there is a connection between utterances $u_i$ and $u_j$, otherwise $e^S_{i,j}=0$. 
Since the speaker relation is bidirectional, we add $e^S_{i,j}=1$ and $e^S_{j,i}=1$ to ${E}^S$ if $u_i$ and $u_j$ are from the same speaker.
Finally, we complete the graph to yield speaker-utterance structures.

\begin{figure}[!t]
  \centering
  \includegraphics[width=0.99\columnwidth]{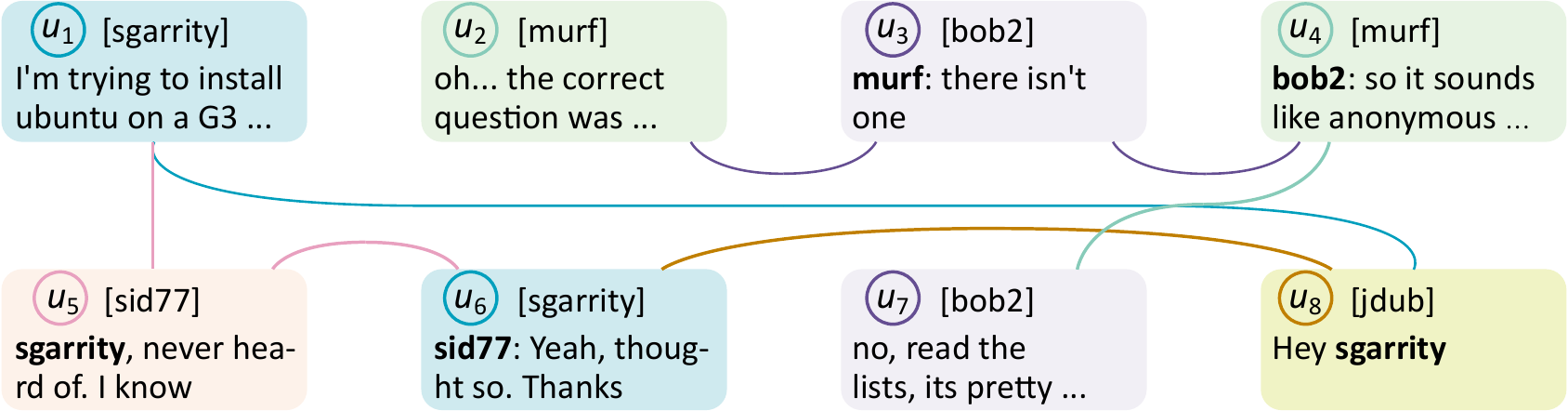}
  \caption{
Examples of the speaker-mention structures, where each utterance that mentions a speaker is connected to the utterance made by that speaker.
  }
  \label{fig:spk_mention}
\end{figure}

\begin{figure}[!t]
  \centering
  \includegraphics[width=0.69\columnwidth]{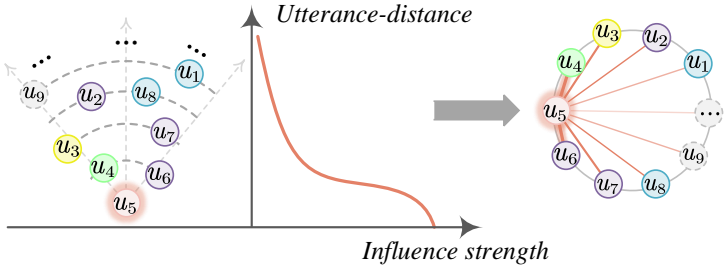}
  \caption{
    Illustrating the utterance-distance structure.
    The edge strength decays by the distance, following the Gaussian distribution.
  }
  \label{fig:gaussion}
\end{figure}

\paratitle{Speaker-mentioning Structure.}
Within multi-party dialogues, it is a commonplace for speakers to reference other participants in their responses, as demonstrated in Figure \ref{fig:spk_mention}.
This speaker co-reference information carries crucial indicators of the participants' roles within the conversation.
Therefore, enhancing speaker-mentioning relationships will contribute significantly to the comprehension of replying relationships and dialogue structures.
To establish these relationships, we create speaker-mentioning edges between the corresponding utterances.
For example, in Figure~\ref{fig:spk_mention}, we establish a connection between $u_1$ and $u_5$ since $u_5$:``\textit{sgarrity, never heard of. I know}'' refers to \textbf{sgarrity}, the originator of $u_1$.
Likewise, we connect $u_4$ with $u_3$ and $u_7$ individually, where the speaker of $u_3$ and $u_7$ is \textbf{bob2}, who is mentioned within $u_4$.
Formally, if $u_i$ or $u_j$ mentions the speaker $s_j$ or $s_i$, the edge $e^M_{i,j}$=1 and $e^M_{j,i}$=1 is established in ${E}^M$. Otherwise, $e^M_{i,j}$ and $e^M_{j,i}$ is set to 0.
This process leads to the creation of the speaker mention graph.

\paratitle{Utterance-distance Structure.}
The context crucial for determining reply dependencies is typically interspersed around the current utterance, with a positive correlation between the degree of dependency and the distance between the two utterances.
This suggests a relational dynamic where closer utterances have a higher likelihood of forming reply-to pairs, while the influence between utterances weakens as the distance between them increases.
To model this tendency, we introduce the concept of utterance-distance, which is characterized using a Gaussian prior distribution.
As depicted in Figure~\ref{fig:gaussion}, utterances within shorter ranges are assigned with higher edge weights, while those at longer distances receive the converse treatment, being assigned lower weights.

The formulation for the initial relationship representation is as follows:
\begin{equation}
\begin{aligned}
\label{eq:basic_dis}
e^{soft}_{i,j} &= \frac{\exp{(\text{Biaf}(\bm{h}_i, \bm{h}_j))})}{\sum_k\exp{(\text{Biaf}(\bm{h}_i, \bm{h}_k))}}, \\
 \text{Biaf} (\bm{h}_i,\bm{h}_j) &=
 \begin{bsmallmatrix}\bm{h}_i \\ 1\end{bsmallmatrix}^T \, \bm{W}^d  \, \begin{bsmallmatrix}\bm{h}_j \\ 1\end{bsmallmatrix} \,,
\end{aligned}
\end{equation}

Following~\cite{gtguo-aaai-2023}, we set the $\mu=0, \sigma=1/\sqrt{2\pi}$ for a Gaussian distribution $\mathcal{N}(\mu,\sigma^2)$ to obtain a distance-aware weight $f(d)=\exp(-\pi d^2)$.
Then we introduce this weight into Eq.~\eqref{eq:basic_dis} to obtain the distance-aware weight:

\begin{equation}
\begin{aligned}
e^{ajs}_{i,j} &= e^{soft}_{i,j}\cdot f(d_{i,j})\\ 
&= \frac{\exp(-\pi(i-j)^2) \cdot \exp(\text{Biaf}(\bm{h}_i, \bm{h}_j)) }{Z_1} \\ 
&= \frac{\exp(-\pi(i-j)^2 + \text{Biaf}(\bm{h}_i, \bm{h}_j)) }{Z_1} \\ 
&=\text{Softmax}_j(\bm{G}(i)),
\end{aligned}
\end{equation}

where $Z_1=\sum_j\exp{\bm{G}(i)_j}$ with $\bm{G}(i)$ being a vector and $\bm{G}(i)_j=-\pi (i-j)^2 + \text{Biaf}(\bm{h}_i,\bm{h}_j)$.
To further refine the equation and balance the two items, we divide them by scale factors $2\pi\sqrt{2\pi}$ and $\sqrt{|i-j|}$, respectively, following the operation in the Transformer~\cite{ash-nips-2017} block.
Thus, we obtain the revised presentation of $e^{ajs}_{i,j}$:

\begin{equation}
\begin{aligned}
\bm{G}'(i)_j & = \frac{(i-j)^2}{2\sqrt{2\pi}} + \frac{\text{Biaf}(\bm{h}_i,\bm{h}_j)}{\sqrt{|i-j|}}\,, \\
e^D_{i,j} & = \text{Softmax}_j(\bm{G}'(i))\,,
\end{aligned}
\end{equation}

where $e^D_{i,j}$ is the distance-aware weight between $u_i$ and other neighbor utterance $u_j$ ($j\neq i$).
As the distance between the $i$-th and $j$-th utterances increases, the weight $e^D_{i,j}$ gradually decreases, resulting in a reduced interaction between them.

\begin{figure}[!t]
  \centering
  \includegraphics[width=0.99\columnwidth]{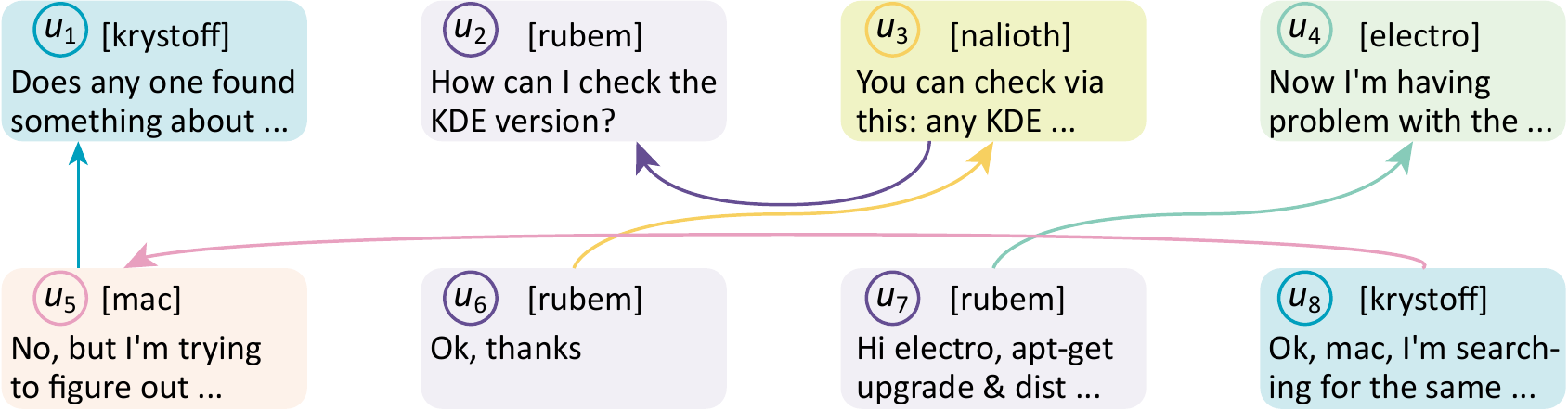}
  \caption{
  Examples of the replying structures, where each pair of utterances has a replying relation, are connected.
  }
  \label{fig:spk_reply}
\end{figure}

\paratitle{Partial-replying Structure.}
As revealed above, the previously discovered partial reply-to structure can actually serve as an important dynamic discourse feature for the following detection.
For example, in Figure~\ref{fig:spk_reply}, \textbf{nalioth} addresses \textbf{rubem}'s question from $u_2$ in $u_3$ by saying ``\textit{You can check via this: any KDE …}''. 
In response, \textbf{rubem} is likely to express gratitude to \textbf{nalioth} in $u_6$.
In essence, if speaker A responds to speaker B in an utterance, then speaker A may reply to B to sustain the dialogue.
Consequently, we construct the partial-replying graph by connecting utterances involved in reply relations with edges $e_{i,j}^R$=1 and $e_{j, i}^R$=1 ($e_{i,j}^R \in {E}^R$).
During the training process, we incorporate all established reply-to relations between utterances to help to optimize the prediction process.
During the inference phase, the reply-to relation is not pre-determined.
In light of this, it's noteworthy that this particular structure $G^R$ undergoes incremental updates as the task decoding progresses, ensuring the constant adaptation and refinement of the model.


\section{Structure-aware Framework}
\label{Structure-aware Framework}
\begin{figure}[!t]
\centering
  \includegraphics[width=0.95\linewidth]{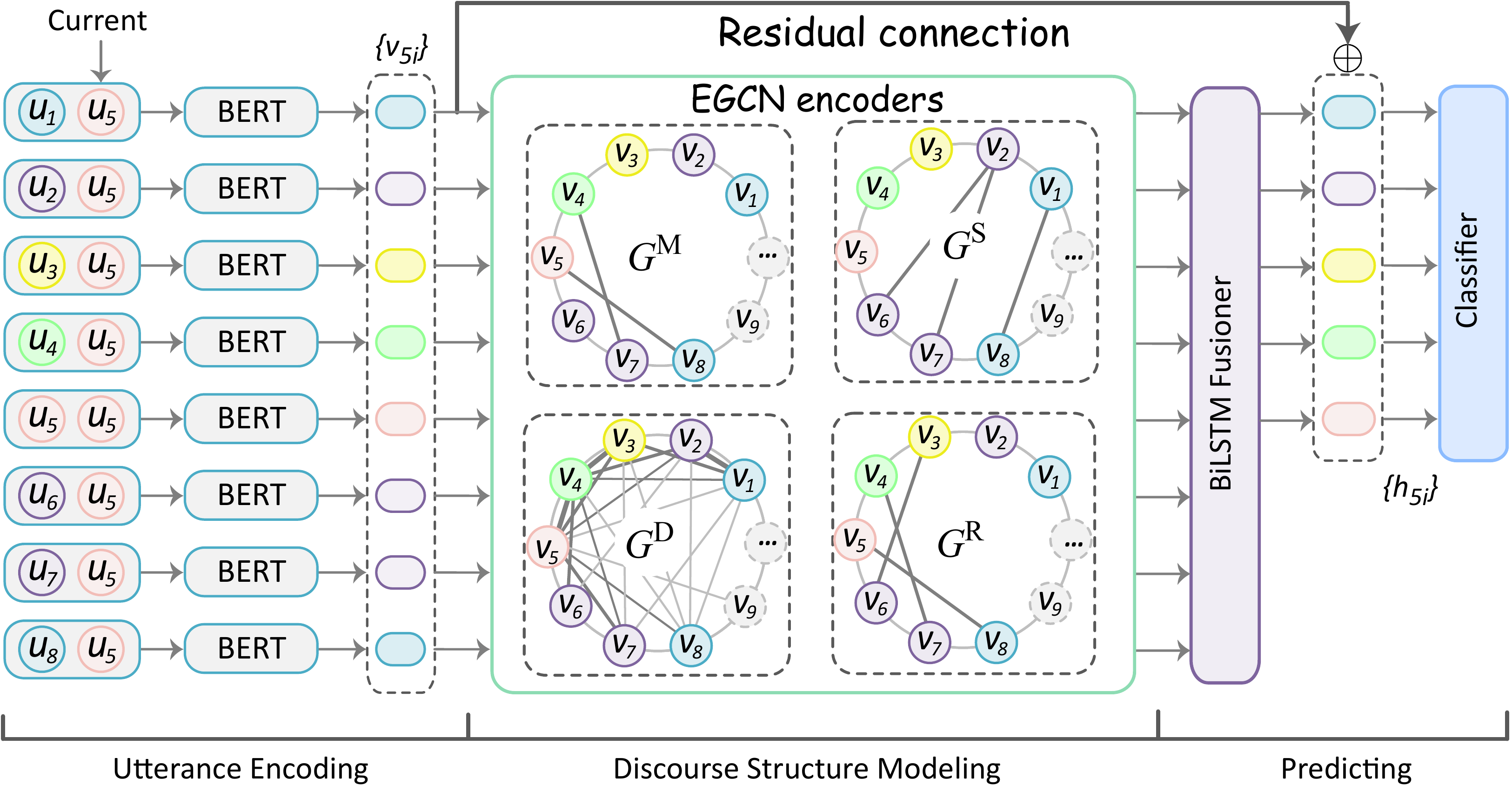}
\caption{
Overview of the structure-aware framework.
The model first accepts utterance pairs (current one, $u_5$, and its candidate parents) as input, and uses BERT for contextual representation encoding.
Then, four types of heterogeneous graphs are encoded by EGCN models to further learn the dialogue-specific discourse structure features.
Based on the final discourse-enhanced contextual features, a classifier decides the parent for the current input utterance.
}
\label{fig:model}
\end{figure}

In this section, we elaborate on the proposed structure-aware framework for dialogue disentanglement, which leverages rich conversational discourse information.
Overall, the architecture of our model consists of three tiers, i.e., \textbf{Utterance Encoding}, \textbf{Discourse Structure Modeling}, \textbf{Predicting}, as illustrated in Figure \ref{fig:model}.

\subsection{Utterance Encoding}
Initially, we convert the input utterances into contextual representations. 
This conversion employs a pair-wise approach to utterance encoding.
Assuming the current utterance is denoted as $u_c$, in an attempt to better understand the context surrounding $u_c$, we incorporate both the preceding and succeeding $\omega$ utterances into its context window.
In other words, we create a dialogue clip, $\overline{U}$, comprising ${u_{c-\omega}, \cdots, u_c, \cdots, u_{c+\omega}}$.
Subsequently, we formulate utterance pair text as follows:
\begin{equation}
X_i = \text{[CLS]}, u_i,\text{[SEP]},u_c,\text{[SEP]},
\end{equation}
where $u_{i}$ denotes a potential parent utterance of $u_c$ within the defined context window, $\overline{U}$.
The special token [CLS] symbolizes the overarching conversation, while [SEP] is employed to demarcate individual utterance pairs.
In the role of a dialogue encoder, we utilize a Pretrained Language Model (PLM).
At the heart of the PLM is a multi-head self-attention block, which calculates its weight via the following formula:
\begin{equation}
\begin{aligned}
 head_i & = \text{Softmax}(\frac{(\bm{QW}_i)(\bm{K}^T\bm{W}_i)}{\sqrt{d}})(\bm{VW}_i), \\
 \text{MultiHeadAtt}(\bm{Q}, \bm{K}, \bm{V}) & = \text{Concat}(head_1,head_2,\cdots,head_h)\bm{W}_o\,.
\end{aligned}
\end{equation}

In this equation, the parameters $\bm{Q},\bm{K},\bm{V}$ are equivalent to the input $\bm{x}$.
Through stacking multiple multi-head self-attention blocks, the PLM is able to uncover and capture intricate contextual dependencies and semantic relationships within the input text.
Therefore, we can derive the token-level representation through the PLM:

\begin{equation}
\begin{aligned}
\{\bm{h}_{CLS}, \bm{h}_1, \cdots, \bm{h}_n\} &=  \text{PLM}( X_i ) ,\\
\bm{v}_{ci} &=  \bm{h}_{CLS}\,,
\end{aligned}
\end{equation}
where $\bm{v}_{ci} \in \mathbb{R}^{d}$ signifies the representation of the utterance pair $\{u_i,u_c\}$, i.e., the representation of $u_c$ in relation to $u_i$.

\subsection{Discourse Structure Modeling}


Building upon the utterance representations, we proceed to encode the discourse structure features as delineated in Section $\S$\ref{Construction of Dialogue Discourse Structures}.
Our inspiration for this step draws from the spatial graph convolutional network (GCN) \cite{kipfW17-iclr,marcheggiani-titov-2017-encoding,wuPCLZY-2021-tnn, ghodag-EMNLP-2019}.
Consequently, we have designed an edge-aware GCN (EGCN), capable of modeling both the topological properties and edge labels of the graph.
In the $l$-th layer of the EGCN, the input for the $i$-th node comprises node features $\bm{r}^t_{ci}(l)$ and an adjacency link set ${e^t_{i,c-\omega},e^t_{i,c-\omega+1},\cdots, e^t_{i, c+\omega} }$, where $t\in{M,S,D,R}$ represents the graph type.
In this scenario, the initial layer node feature $\bm{r}^t_{ci}(0)$ is initialized using the representation $v_{ci}$ procured from the PLM.
The adjacency link $e^t_{i,j}$ is derived from the corresponding structure.
As per the definitions in Section~\ref{Construction of Dialogue Discourse Structures}, the value of $e^t_{i,j}$ in speaker-utterance structure, speaker-mentioning structure, and partial-replying structure is set to 1 if a link exists, otherwise it is set to 0.
In the case of the utterance-distance structure, $e^t_{i,j}$ is a continuous value that changes in accordance with the value of $|i-j|$.
Finally, the output of the $l$-th layer for the $i$-th node is the updated feature $\bm{r}^t_{ci}(l+1)$.

Notably, we utilize four distinct EGCNs for graph modeling, each corresponding to one of the four heterogeneous graphs ($G^M$, $G^S$, $G^D$, and $G^R$).
The EGCN operation can be formalized as follows:
\begin{equation}
\begin{aligned}
\label{EGCN-1}
 \bm{r}^{t}_{ci}(l+1) &= \text{Sigmoid}(
\sum_{j=c-\omega}^{c+\omega} \pi^{t}_{i,j}(l)(\bm{W}^t(l) 
\bm{r}_{cat}^{t,c,i,l}
+ \bm{b}^t(l))
) \,, \\
\bm{r}_{cat}^{t,c,i,l} &= [\bm{r}^t_{ci}(l) ; \bm{r}^t_{cj}(l) ; \bm{a}^{t} ] \,.
\end{aligned}
\end{equation}
In this equation, $\bm{W}^t(l)$ and $\bm{b}^t(l)$ stand for the trainable parameters for the $l$-th layer.
The edge label embedding of $t$ type of graph is represented by $\bm{a}^{t} \in \mathbb{R}^{a}$, and it is randomly initialized.
EGCN weight $\pi^{t}_{i,j}(l)$ guides the graph aggregation:
\begin{align}
\label{EGCN-2}
\pi^{t}_{i,j}(l) &= \frac{
e^{t}_{i,j} \cdot \text{Agg}(i,j,t,l)
}
{
\sum\limits_{k=c-\omega}^{c+\omega} e^{t}_{i,k} \cdot \text{Agg}(i,k,t,l)
} \,, \\
\text{Agg}(i,j,t,l) &= \exp{(\text{Biaf}([\bm{r}^t_{ci}(l); \bm{a}^{t}], [\bm{r}^t_{cj}(l);\bm{a}^{t}])} \,,
\end{align}
where $\text{Biaf}(\cdot, \cdot)$ is the Biaffine function defined in Eq.~\eqref{eq:basic_dis}.
In this case, $\text{Biaf}(\cdot, \cdot)$ refers to the Biaffine function defined in Eq.~\eqref{eq:basic_dis}.
It should be noted that we implement a total of $L$ layers of EGCN propagation for each graph.
This is done to ensure comprehensive learning of structural features. The final representation for each node's corresponding discourse structure is denoted by the output of the last layer $\bm{r}^t_{ci}(L)$.
Following this, we propose fusing the heterogeneous graph features into a unified representation by performing an addition operation:
\begin{equation}
 \bm{r}_{ci} =  \bm{r}^{M}_{ci}(L) \oplus  \bm{r}^{S}_{ci}(L) \oplus  \bm{r}^{D}_{ci}(L) \oplus  \bm{r}^{R}_{ci}(L) \,.
\end{equation}


\subsection{Predicting}

In addition to the non-linear discourse structure, the natural chronological order also provides vital cues for understanding the entirety of the dialogue.
To effectively exploit the temporal dependencies embedded in dialogues, we utilize a BiLSTM model as the foundational component of our methodology~\cite{hoclstm-nc-1997}.
Given the overfitting issues prevalent in the conventional LSTM model~\cite{saklsm-inter-2014}, we incorporate a DropoutConnect BiLSTM~\cite{ilyoti-icml-2013, MerityKS18-iclr} to more efficiently extract temporal dependencies.
The following equation succinctly illustrates the underlying mechanism of this model:

\begin{equation}
\begin{aligned}
\bm{i}_t &= \sigma(\bm{W}_i\bm{x}_t + (\bm{M}_i \cdot \bm{U}_i) \bm{h}_{t-1} + \bm{b}_i)\,, \\
\bm{f}_t &= \sigma(\bm{W}_f\bm{x}_t + (\bm{M}_f \cdot \bm{U}_f) \bm{h}_{t-1} + \bm{b}_f)\,, \\
\bm{o}_t &= \sigma(\bm{W}_o\bm{x}_t + (\bm{M}_o \cdot \bm{U}_o) \bm{h}_{t-1} + \bm{b}_o)\,, \\
\tilde{\bm{c}}_t &= \tanh(\bm{W}_{c}x_t + (\bm{M}_c \cdot \bm{U}_c) \bm{h}_{t-1} + \bm{b}_c)\,, \\
\bm{c}_t &= \bm{f}_t \odot \bm{c}_{t-1} + i_t \odot \tilde{\bm{c}}_t\,, \\
\bm{h}_t &= \bm{o}_t \odot \tanh(\bm{c}_t)\,,
\end{aligned}
\end{equation}
where the parameters $[\bm{W}_*, \bm{U}_*, \bm{b}_*]$ are trainable, with $* \in \{i,f,o,c\}$. 
$\bm{M}_*$, a binary matrix, signifies the dropout connection and importantly, $\bm{M}*^{ij}$ follows a Bernoulli distribution with a probability parameter $p$.
Applying Drop-Connect to the hidden-to-hidden weight matrices $[\bm{U}_i, \bm{U}_f, \bm{U}_o, \bm{U}_c]$ can regularize the models sufficiently, thereby overfitting on the recurrent connections of the LSTM can be effectively mitigated~\cite{MerityKS18-iclr}.
For each time step, we adopt the node's representation from the Edge-aware GCN (EGCN), denoted as $\bm{r}_{ci}$, as the input $\bm{x}_t$.
The hidden state at each time step in the final layer, denoted as $\bm{h}_{ci}$, serves as the representation of the $i$-th node. 

Then, our model architecture incorporates a residual connection between the initial utterance representation $\bm{v}{ci}$, derived from the PLM, and the current discourse feature $\bm{h}_{ci}$.
This serves as the final contextual feature for the subsequent prediction:
\begin{equation}
\label{eq:concat}
 \hat{\bm{h}}_{ci} =  \bm{h}_{ci} \oplus  \bm{v}_{ci} \,,
\end{equation}
where $\bm{h}_{ci}$ represents the output from the last layer of the BiLSTM.

Building upon this final discourse-enhanced contextual feature representation, a feedforward neural network (FFNN) is utilized to compute a pairwise score:
\begin{equation}
\label{eq:score}
S(ci) = FFNN( \hat{\bm{h}}_{ci} ) \,,
\end{equation}
Here, $S(ci)$ serves as a unary score that signifies the confidence level that $u_c$ responds to $u_i$.
Subsequently, a softmax layer is applied across all candidates to identify the final parent of $u_c$:
\begin{equation}
\label{eq:classifer}
p_{ci} = \frac{\exp(S(ci))}{\sum_{c-\omega <j\leq c}\exp(S(cj))} \,.
\end{equation}
The candidate with the highest $p_{ci}$ is identified as the parent.
It is critical to remember that the dialogues are arranged in chronological order, so a potential parent can only precede $u_c$ or be $u_c$ itself,
\footnote{
Please note that a node $u_c$ can be its parent itself when $u_c$ is the start utterance of a session.} 
i.e., $c-\omega<i\le c$.

\section{Framework Training and Inference}

In this section, we elaborate on how the framework is training on the train set (cf $\S$\ref{sec:hi-ranking}), and then how to make predictions on the test set (cf $\S$\ref{sec:Easy-First Graph-based Decoding}).
Both two stages are optimized by fully considering the conversational discourse characteristic.

\subsection{Learning with Hierarchical Ranking Loss}
\label{sec:hi-ranking}

The evaluation of dialogue disentanglement not only measures the pair-wise detection of the reply-to relation but also considers the correctness of the session cluster-level prediction.
Considering a case where the parent of current utterance $u_c$ is wrongly assigned as $u_k$ that belongs to the same session with $u_c$, the prediction will be considered valid under the cluster-level perspective, even $u_k$ is not $u_c$'s parent.
Therefore, different from regular learning tasks, if only minimizing the cross-entropy loss over gold replying relational pairs, the model will be guided to ignore the session-level learning and result in biased prediction.
To this end, we propose a hierarchical ranking loss (HRL) for model optimization.

\begin{figure}[!t]
\centering
  \includegraphics[width=0.82\columnwidth]{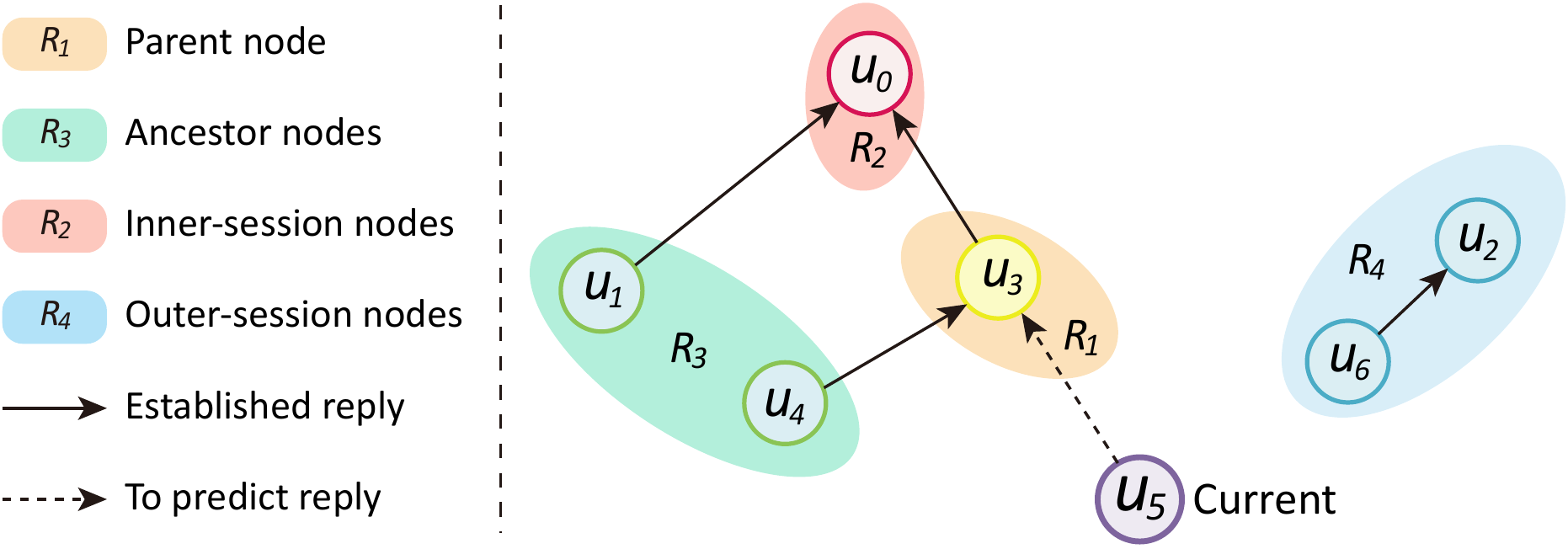}
\caption{
Dividing utterance's parent candidates into four levels.
Here $u_5$ is the current utterance, and $u_3$ is the golden parent of $u_5$.
Under cluster-level evaluation, it is regarded as correct even if $u_0, u_1$ or $u_4$ is wrongly predicted as $u_5$'s parent.
}
\label{fig:setwise-ranking}
\end{figure}

Specifically, we consider the training of our system under multiple levels.
Given the current utterance $u_c$, and currently other utterances $\overline{U}$ with established replying relations, we now need to determine the replying relation from $u_c$ to any candidate in $\overline{U}$.
Considering that dialogues are characterized by intrinsic discourse with hierarchical structures, we can naturally divide all $u_c$'s parent candidates
into four different levels, as illustrated in Figure \ref{fig:setwise-ranking}:
\begin{compactitem}
    \item Parent utterance node ($R_1$),
    which is the first-order and only parent of $u_c$.
    \item Ancestor utterance nodes ($R_2$), where the utterances are under the same thread, having a second-order relation with $u_c$ in the two-hop distance.
    \item Inner-session utterance nodes ($R_3$), where utterances belong to the same session with $u_c$ but these utterances are not included in $R_1$ or $R_2$.
   \item Outer-session utterance nodes ($R_4$), where utterances are separated in different conversational sessions with $u_c$.
\end{compactitem}

Based on the above division of utterances, we can denote the training objectives of HRL into three different levels:
\begin{align}
\mathcal{L}_{1} &= -\log{\frac{\sum_{u_i \in R_1}{\exp{S(ci)} }}{ \sum_{u_j \in R_{1}\cup R_{2}\cup R_{3}\cup R_{4}}{\exp{S(cj)}}}} \,, \\
\mathcal{L}_{2} &= -\log{\frac{\sum_{u_i \in R_2}{\exp{S(ci)} }}{ \sum_{u_j \in R_{2}\cup R_{3}\cup R_{4}}{\exp{S(cj)}}}}  \,,\\
\mathcal{L}_3 &= -\log{\frac{\sum_{u_i \in R_3}{\exp{S(ci)} }}{ \sum_{u_j \in R_{3}\cup R_{4}}{\exp{S(cj)}}}}  \,,
\end{align}
where $\mathcal{L}_{1}$ indicates the regular pair-wise optimization; $\mathcal{L}_{2}$ and $\mathcal{L}_{3}$ aim at the session-wise optimizations.
$S(ci)$ is the unary score obtained via Eq.~\eqref{eq:score}.

We can summarize the final total loss as follows:
\begin{equation}
\mathcal{L} = \alpha_1 \mathcal{L}_1 +  \alpha_2 \mathcal{L}_2 +  \alpha_3 \mathcal{L}_3 \,.
\end{equation}
where $\alpha_{*} \in (0, 1]$ are the weighting coefficients.

\begin{figure}[!t]
\centering
\includegraphics[width=0.78\columnwidth]{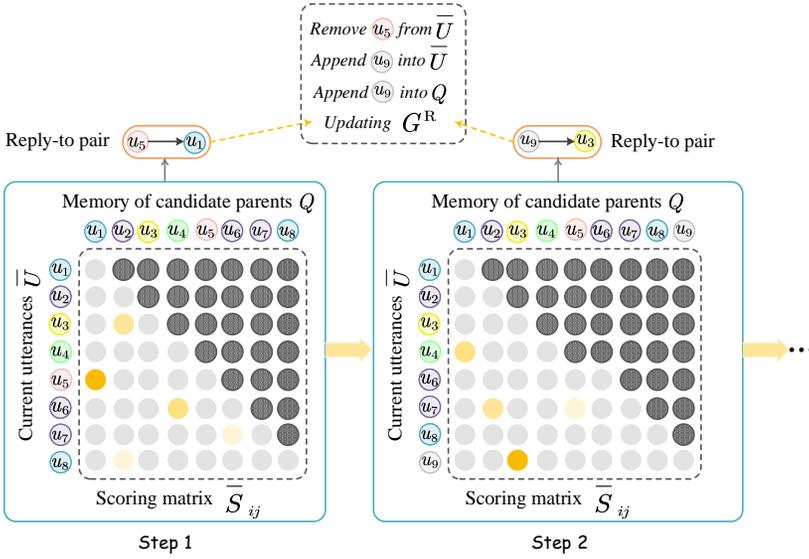}
\caption{
Illustration of the easy-first decoding.
}
\label{fig:easy-first}
\end{figure}

\subsection{Easy-first Replying Relation Decoding}
\label{sec:Easy-First Graph-based Decoding}

In the inference stage, all the existing work decodes the relying relation for each utterance from the front-to-back direction.
A clear drawback of such practice is the local and shortsighted context modeling: the decision at each step is made with the prior information.
Also, the conversation utterances are essentially organized with structured discourse, and thus it is not sensible to take a sequential decoding order.
Here we mimic the human-like inference of the replying relation, executing the decoding by looking at the global contexts in a non-directional manner, i.e., with an easy-first decoding algorithm.
We start from easy determining decisions and proceed to harder and harder ones.
When making later decisions, the system has access to the entire structure built in earlier stages.

We first build a pair-wise utterance graph over two utterance lists to implement the idea: the current utterance set $\overline{U}$ and the memory of candidate parents $Q$.
$Q$ caches the utterances in the current dialogue $U$, and any utterance $u_i$ in $\overline{U}$ will select its potential parent utterance from $Q$.
Initially, ${Q}$ is loaded with $\overline{U}$.
By Cartesian multiplication over $\overline{U}$ and $Q$, we maintain a scoring matrix (i.e., utterance graph):
\begin{gather}
\label{scoring}
\overline{S}_{ij}  = \begin{cases}
{S}(ij), \quad & i - \omega < j \leq i \\
-\infty,  \quad & otherwise
\end{cases}
\end{gather}
 where
 $S(ij)$ is the pair-wise score calculated via Eq. \eqref{eq:score}, which is set to $-\infty$ when $j$ is out of the range $(i-\omega, i]$, such that the parent node of $u_i \in \overline{U}$ is always the antecedent utterance or itself within a certain range $\omega$.
For each scoring iteration, we select the pair ($u_i^{\curvearrowright}u_j$) with the highest score, i.e., the most confident pair, as a valid replying pair in the current step.

Thereafter, we add the pair $u_i^{\curvearrowright}u_j$ into the partial-replying graph $G^R$ and update it.
Also, we remove $u_i$ from $\overline{U}$, as $u_i$ has already found its parent; and append the successor utterance $u_{1+i+\omega}$ into the window as new $\overline{U}$.
Meanwhile, we maintain $Q$ by adding the following utterance $u_{1+i+\omega}$ into $Q$ each time, such that each utterance in $\overline{U}$ has the chance to find its parent in $Q$.
The decoding follows the window sliding iteratively, until all elements in $U$ find their parent, and $\overline{U}$ is empty.
In Figure \ref{fig:easy-first}, we depict the easy-first decoding, which is formulated by the Algorithm in Appendix $\S$\ref{alg:decoding}.

\begin{algorithm}[!h]
\caption{Easy-first Decoding for Replying Relation}
\label{alg:decoding}
\SetKwInput{KwInput}{Input}                
\SetKwInput{KwOutput}{Output}              
\DontPrintSemicolon

  \KwInput{Current dialogue queue $\overline{U}$;\\
  Initiate $Q \gets \overline{U}$; \\
  Initiate partial-replying Graph $G^R$=$\{{V}$, $\diameter\}$.
  }

  \SetKwFunction{FMain}{Main}
  \SetKwFunction{FSum}{Easy-First}
  \SetKwFunction{FInf}{Inference}
  \SetKwFunction{FSub}{Sub}
  \SetKw{KwBy}{by}

\While{ $\overline{U}$ \text{is not empty} }{

    $\overline{S}$= Pairwise-Scoring($\overline{U}, Q$ | $\{\bm{h}_{ci}\}$); \tcp*{Get the pair-wise scores under current phase via Eq. \ref{scoring}}

    $ i_s, j_p = \mathop{\arg \max}\limits_{i,j} {\overline{S}}$; \tcp*{Search for the highest-scored parent index ($j_p$) from $Q$ and the child index ($i_s$) from $\overline{U}$}

    $G^R$.Update($u_{i_s}^{\curvearrowright}u_{j_p}$); \tcp*{Add the established pair into $G^R$}

    Pop $u_{i_s}$ from $\overline{U}$; \tcp*{Update $\overline{U}$}
    Append $u_{1+i_s+\omega}$ in the rear of $\overline{U}$; \tcp*{Update $\overline{U}$}

    Append $u_{1+i_s+\omega}$ in the rear of $Q$; \tcp*{Update $Q$}

}
  \KwOutput{${G}^R$=$({V},{E}^R)$}
\end{algorithm}

\textbf{$\bullet$ Remark.}
It is noteworthy that during training, with gold replying annotations as training signals, the model phase does not need to take the easy-first decoding.\footnote{
Meanwhile, the training set comes without annotations for the easy-first procedure.
}
Also, in the training stage, the gold replying data enables the partial-replying structure $G^R$ always to be of high-quality to guide the learning of further replying relations.
During inference, easy-first decoding is engaged to give better inference, as cast above.
However, at inference time, the $G^R$ we used is based on the previous model predictions, where the errors can be accumulated to worsen the following prediction.
To minimize the gap between training and inference, during the training phase, we purposely create some noises by randomly replacing the gold parent of an utterance with the wrong utterance, at a probability of 15\%.
Such a teacher-forcing strategy helps enhance the model generalization ability and maintain stable training.

Algorithm \ref{alg:decoding} summarizes the easy-first inference algorithm in a formal format.

\section{Experiment Settings}

\subsection{Dataset}


\begin{table}[!h]
\fontsize{9.5}{11.5}\selectfont
 \setlength{\tabcolsep}{4mm}
    \centering
        \caption{
Statistics for Ubuntu IRC and Movie Dialogue dataset.
}
    \begin{tabular}{ll rrrr}
    \hline
    \bf Dataset & \bf Item & \bf Train & \bf Valid & \bf Test & \bf Total \\
    \hline
   \multirow{8}{*}{\bf Ubuntu IRC}  & Dialogue & 153 & 10 & 10 & 173 \\
    & Session   & 17,267 & 494 & 961 & 18,722 \\
    & Utterance & 67,463 & 2,500  & 5,000 & 74,963 \\
    & Speaker   &  33,743 &  1,953 & 2,469 & 42,165  \\
    & Char / Utterance &  75.58 &  73.10 & 77.76 & 75.58 \\
    & Word / Utterance &  32.86 &  31.83 & 33.67 & 32.86 \\
    & Utterance / Session &  12.15 & 11.21  & 14.08 & 12.22 \\
    & Utterance / Dialogue &  440.93 &  250 & 500 & 433.31 \\
    & Session / Dialogue &  10.03 & 9.99  & 12.38 & 10.17 \\
    \hline
   \multirow{8}{*}{\bf Movie Dialogue}  & Dialogue & 29,669 & 2,036 & 2,010 & 33,715\\
    & Session   & 89,064 & 6,057 & 6,058 & 101,179 \\
    & Utterance & 728,043 & 49,644  & 49,506 & 827,193 \\
    & Speaker   &  268,014 &  18,473 & 18,362 & 304,849  \\
    & Char / Utterance & 60.17 & 60.24 & 59.71 & 60.16 \\
    & Word / Utterance & 27.75 & 27.78 & 27.65 & 27.74 \\
    & Utterance / Session &  8.17 &  8.20 & 8.17  & 8.18 \\
    & Utterance / Dialogue & 24.54 & 24.38 & 24.62 & 24.53 \\
    & Session / Dialogue & 3.00 & 2.97 & 3.01 & 3.00 \\
    \hline
    \end{tabular}
\label{table:corpus_statistic}
\end{table}

Our empirical analysis is conducted on two publicly available dialogue disentanglement datasets, namely, the Ubuntu IRC corpus and the Movie Dialogue dataset.

The Ubuntu IRC dataset~\cite{kummerfeld-etal-2019-large} is a rich corpus that provides a wealth of question-and-answer content related to the Linux operating system. This dataset is characterized by its depth and complexity, containing 173 conversations that collectively comprise a staggering 74,963 utterances. The average dialogue in this dataset is notably extensive, with an average of 433.3 utterances per conversation. This characteristic makes the Ubuntu IRC dataset a challenging and valuable resource for dialogue disentanglement research, as it offers a diverse range of conversational threads and interactions to analyze and understand.

The Movie Dialogue dataset~\cite{liuSGLWZ20-ijcai}, is a unique resource that has been annotated based on movie scripts. This dataset is composed of 33,715 conversations, providing a substantial volume of dialogue for analysis. However, in contrast to the Ubuntu IRC dataset, the conversations in the Movie Dialogue dataset are typically shorter, averaging 24 utterances per dialogue. This dataset was developed by extracting 56,562 sessions from 869 movie scripts that explicitly indicate plot changes. These sessions were then manually intermingled to create a synthetic dataset, with the minimum and maximum session numbers in one dialogue being 2 and 4, respectively. The Movie Dialogue dataset offers a different kind of challenge for dialogue disentanglement, as it involves understanding and separating intertwined dialogues from the context of movie scripts.

These two datasets, with their distinct characteristics and challenges, provide a comprehensive basis for our experiments in dialogue disentanglement. The Ubuntu IRC dataset, with its extensive and complex dialogues, and the Movie Dialogue dataset, with its intertwined dialogues from movie scripts, together offer a broad and diverse range of data for our research.
Table~\ref{table:corpus_statistic} shows the detailed data statistics.

\begin{table}[!t]
\fontsize{9.5}{10.5}\selectfont
\setlength{\tabcolsep}{5.mm}
    \caption{
    Hyperparameters setting.
    }\label{param-del}
    \centering
    \begin{tabular}{llll}
    \hline
    \bf Parameter/Module & \bf Value\\
    \hline
    \bf BERT & \\
        \quad version & bert-base-uncased & \\
        \quad hidden size & 768 & \\
    \hline
    \bf E-GCN & \\
        \quad layer num & 2 & \\
        \quad hidden size & 768 & \\
        \quad dropout & 0.2 \\
        \quad label embdding dim& 10 \\
        \quad activate function & relu~\cite{nairH-2010-icml} & \\
    \hline
    \bf LSTM &   \\
        \quad version & WeightDrop~\cite{MerityKS18-iclr} \\
        \quad layer num & 2 & \\
        \quad hidden size & 768 & \\
        \quad feedforward dropout & 0.2 \\
        \quad recurrent dropout & 0.2 \\
        \quad bidirectional & True \\
    \hline
    \bf FFNN & \\
        \quad layer num & 2 \\
        \quad hidden size$_{layer1}$ & 768 & \\
        \quad hidden size$_{layer2}$ & 300& \\
        \quad dropout & 0.2 & \\
        \quad activate function & relu~\cite{nairH-2010-icml} & \\
        \hline
    \bf Training & \\
        \quad learning rate (bert layer)& $6e^{-6}$ \\
        \quad learning rate (non-bert layer)& $1e^{-5}$ \\
        \quad optimizer & AdamW \\
        \quad batch size & 2 \\
        \quad epoch size & 4 \\
    \cdashline{1-4}
    \hline
    \end{tabular}
\end{table}

\subsection{Implementations}

\paratitle{Hyper-parameters} In this study, we utilize BERT-base (uncased)\footnote{\url{https://huggingface.co/bert-base-uncased}} as our backbone PLM.
To further improve the performance of our model, we apply a dropout~\cite{SrivastavaHKSS14-jmlr} layer with a rate of 0.2 after the encoder.
We set batch size as two dialogues.
The initial learning rate for BERT and non-BERT layers are set to 6e-6 and 1e-5, respectively.
AdamW optimizer \cite{loshchilovH2019-iclr} and LR scheduler with warmup mechanism \cite{GoyalDGNWKTJH17-arxiv} are employed for optimization.
We use a two-layer EGCN with a 768-d hidden size.
The FFNN in Eq.~\eqref{eq:score} is with two layers and 300-d.
$\alpha_1,\alpha_2, \alpha_3$ is set to 1, 0.1, and 0.05, respectively, based on the development experiment.
$\omega$ is set to 50, and the maximum length of the utterance pair is limited to 128.
All experiments were conducted on Ubuntu systems with two RTX 3090 GPUs.
In Table \ref{param-del}, we show the detailed parameter choosing.

\paratitle{Details of model} In the data processing stage, no special processing is applied to the text strings.
Instead, the dialogue text is segmented using the tokenizer provided by BERT.
For utterance pairs exceeding the maximum allowed length, truncation is applied.
If the shorter utterance is less than half the length, the longer utterance is truncated to ensure the pair length does not exceed the maximum length.
If two utterances are longer than half the maximum allowed length, both are truncated to half the maximum length to preserve the information as fairly as possible.
During the graph construction phase, the speaker-utterance graph is directly constructed using the given speaker information for each utterance.
The speaker-mentioning graph is constructed by matching the speaker's name of the previous utterance, establishing the speaker-mentioning relationship.


\subsection{Baselines}
We adopt the prior strong-performing dialogue disentanglement systems as our baselines.
Following, we give a brief description of each baseline:

\begin{compactitem}
    \item \textbf{FeedForward} ~\cite{kummerfeld-etal-2019-large} predicts the replying relation with simple feedforward networks.

    \item \textbf{Elsner} ~\cite{elsner-charniak-2008-talking} takes a pairwise classification with handcrafted features for determining the replying relation.

    \item \textbf{BERT} ~\cite{devlin-etal-2019-bert} is the vanilla BERT model, which treats the task as a sentence pair classification problem, i.e., predicting reply-to relationships.

    \item \textbf{BERT+MF} ~\cite{zhu-etal-2021-findings} combines BERT with handcrafted features for dialogue disentanglement.

    \item \textbf{Transition} ~\cite{liuSGLWZ20-ijcai} solves the dialogue disentanglement with a transition-based model.

    \item \textbf{DiaBERT} ~\cite{tiandali-arxiv} employs a hierarchical transformer to enhance thread and session features extraction, during which additional handcrafted pairwise features (e.g., token-level overlapping) are incorporated into the model to enhance performances.

    \item \textbf{Struct}~\cite{aclMa0Z22} constructs an utterance-level graph to strengthen the utilization of discourse structural information, and further leverage refined LSTM to model the context information.

    \item \textbf{PtrNet}~\cite{yu-joty-2020-online} uses the pointer networks for dialogue disentanglement. It selects a parent for the current utterance.

    \item \textbf{Bi-Level}~\cite{chengyuhuang-emnlp} incorporates a bi-level (thread, session) contrastive loss into dialogue disentanglement.
\end{compactitem}

Note that \emph{Struct} is the current best-performing system for the task.
To ensure fair comparisons, most of the recent baselines use the PLM with the base version.
The baseline results are copied from their raw papers, while we re-implement the \emph{Struct} model for some analyzing experiments.

\subsection{Evaluations}
Following prior works \cite{zhu-etal-2021-findings,aclMa0Z22}, we evaluate the task performance with the cluster-level and exact pair-wise metrics.
Also, we add a new metric for in-depth analysis (see Figure~\ref{fig:session_exp}).
Here we provide a detailed description of the evaluation metric used in our experiments.

\paratitle{1) Cluster-level metric:}
\begin{compactitem}
    \item \textbf{Variation of Information} (VI) \cite{meila03-colt} measures the information change between two sessions:
    $VI(X, Y)=2 * H(X, Y) - H(X) - H(Y)$, where $X$ and $Y$ are golden and predicted clusters, and $H(\cdot)$ is the entropy.
    In line with prior research~\cite{kummerfeld-etal-2019-large}, we present $1-VI$ as the final metric, where a higher value indicates better performance.

    \item \textbf{Adjusted Rand Index} (ARI) ~\cite{hubert1985comparing} evaluates the similarity between golden and predicted clusters, by counting all identical pairs in two clusters:
    \begin{equation}\label{eq:ari}\small
        \text{ARI}(X,Y) = \frac{ \sum_{ij}
        {\tbinom{n_{ij}}{2} }
        - [\sum_i {\tbinom { a_i} {2} } \sum_j {\tbinom { b_j} {2} } ] / \tbinom{n}{2}
        }
        {
        \frac{1}{2} [\sum_i {\tbinom { a_i} {2} } \sum_j {\tbinom { b_j} {2} } ]
        - [\sum_i {\tbinom { a_i} {2} } \sum_j {\tbinom { b_j} {2} } ] / \tbinom{n}{2}
        } \,,
    \end{equation}
Where $n_{ij}$=$| X_i \cap Y_j |$ is the amount of common elements between $X_i$ and $Y_j$,
$a_i$=$\sum_j {n_{ij}}$ and $b_j$=$\sum_i {n_{ij}}$ is the row-wise and column-wise sum of $n_{ij}$.

    \item \textbf{One-to-one} (1-1) ~\cite{elsner-charniak-2010-disentangling} reflects the ability to summarize the whole conversation.
    The metric is computed by finding an optimal matching bipartite between two clusters from the golden and prediction set.

    \item \textbf{Normalized Mutual Information} (NMI) ~\cite{aaron-arxiv-nmi} also evaluate cluster-level prediction,
    which calculates the mutual information between two sets:
    \begin{equation}\small
        \text{NMI}(X, Y) = \frac{2\times I(X;Y) } {[H(X) + H(Y)]} \,,
    \end{equation}
    where $I$ is the mutual information.

    \item \textbf{Local$_{k}$} \cite{elsner-charniak-2010-disentangling} measures the accuracy of determining whether an utterance belongs to the same cluster as the previous $k$ utterances:
\begin{equation}\small
\text{Local}_k =\frac{
    \sum_{j=1}^k \sum_i{\bm{1}(
    \bm{1}(C_p^{u_i} = C_p^{u_{i-j}}) = \bm{1}(C_p^{u_i} = C_p^{u_{i-j}}) )}} {\sum_{j=1}^k \sum_i 1} \,,
\end{equation}
where $C_p^{u_i}$ and $C_g^{u_i}$ are the predicted/gold cluster that $u_i$ belongs to, $\bm{1}(x$=$y)$ is a binary indicator returning 1 if $x$=$y$, else 0.
In our practice, we set $k$=$3$.

    \item \textbf{Shen-F1} (S-F)~\cite{sigir-ShenYSC06} assesses how well the predicted clusters align with the golden clusters:
    \begin{equation}\small
        \text{Shen-F} = \sum_i \frac{n_i}{n} \mathop{\max}\limits_{i}
        {\frac{2n_{ij}^2}
        {n_i + n_j}} \,,
    \end{equation}
where notations are same as in Eq.~(\ref{eq:ari}).

\item \textbf{Cluster Exact Match (P, R, F1)}
evaluates the precision (P), recall (R), and F1 (F1) scores at the cluster level, where a cluster is considered a match if completely identical to the golden set.

\end{compactitem}

\paratitle{2) Pair-wise metric:}
\begin{compactitem}
\item \textbf{Link Exact Match (P, R, F1)}
evaluates exact matchings of all reply-to pairs.
\end{compactitem}

\paratitle{3) Additional metric}
The original cluster-level evaluation metrics are calculated on the complete cluster set, which cannot be applied to the subset evaluation.
To evaluate the performance of the subset, we improved the evaluation index based on ARI by proposing a new metric called partial-ARI:
\begin{equation}
\label{eq:pari}
Par\text{-}ARI(\mathcal{S}, Y) = ARI(\mathcal{S}, \{Y_i | Y_i \in Y, \exists \phi \in \mathcal{S}:  |Y_i \cap \phi| > 0\})
\end{equation}
where $Y$ contains the predicted clusters and $\mathcal{S}$ contains gold clusters whose size is in a certain range.
Based on this formula, we can group golden clusters by size and then calculate the score for each group.



The best-performing model on the validation set is used for testing, and the average results of five runs were reported.
All scores from our implementation are after paired t-test with $p<$ 0.05.

\section{Experiment Result and Discussion}

\subsection{Main Results and Observations}


\begin{table*}[!t]
\fontsize{9.5}{12.5}\selectfont
\caption{
Performance on Ubuntu IRC dataset under cluster-level evaluating metrics.
In the brackets are the improvements of our scores over the best baseline (underlined).
}
    \centering
    \begin{tabular}{l ccccccccc}
    \toprule
  \multicolumn{1}{c}{\multirow{2}{*}{\textbf{Methods}}}  & \multicolumn{6}{c}{ \textbf{Set Measuring}} & \multicolumn{3}{c}{\textbf{Exact Matching}} \\
        \cmidrule(r){2-7}\cmidrule(r){8-10}
         & \textbf{VI} & \textbf{ARI} & \textbf{1-1} & \textbf{NMI} & \textbf{Local}$_3$ & \textbf{S-F} & \textbf{P} & \textbf{R} & \textbf{F1}\\ \hline
         FeedForward~\cite{kummerfeld-etal-2019-large} & 91.30 & - & 75.60 & - & - & - & 34.60 & 38.00 & 36.20\\
         Elsner~\cite{elsner-charniak-2008-talking} & 82.10 & - & 51.40 & - & - & - & 12.10 & 21.50 & 15.50\\
         Lowe~\cite{mehri-carenini-2017-chat} & 80.60 & - & 53.70 & - & - & - & 10.80 & 7.60 & 8.90 \\
         BERT~\cite{liuSGLWZ20-ijcai} & 90.80 & 62.90 & 75.00 & - & - & - & 32.50 & 29.30 & 36.60 \\
         BERT+MF~\cite{zhu-etal-2021-findings} & 92.00 & - & 77.00 & - & - & - & - & - & 40.90 \\
         Transition ~\cite{liuSGLWZ20-ijcai} & - & - & - & 62.10 & 20.58 & 49.70 & - & - & - \\
          PtrNet ~\cite{yu-joty-2020-online} & 94.20 & - & 80.10 & - & - & - & \underline{44.90} & 44.20 & 44.50 \\
          DiaBERT~\cite{tiandali-arxiv} & 93.20 & 72.80 & 79.70 & - & - & - &  42.10 & \underline{47.90} & \underline{44.80}  \\
          Struct ~\cite{aclMa0Z22} & \underline{93.30} & \underline{73.60} & \underline{81.18} & \underline{89.04} & \underline{94.08} & \underline{84.64} & 41.46 & 44.23 & 42.90 \\
          \hdashline
        \bf Ours & \textbf{94.23} & \textbf{81.10} & \textbf{84.20} &  \textbf{91.85} & \textbf{95.64} & \textbf{87.50}  &  \textbf{47.97} & \textbf{49.86} & \textbf{48.90} \\
        \specialrule{0em}{-2pt}{-1pt} & \small{(+0.03)}&         \small{(+7.50)}&\small{(+3.02)}&\small{(+2.81)}&\small{(+1.56)}&\small{(+2.86)}&\small{(+3.07)}&\small{(+1.96)}&\small{(+4.10)}\\
        \hline
    \end{tabular}
    \label{table:mainres}
\end{table*}

Table~\ref{table:mainres} shows the comparisons with baselines on the Ubuntu IRC dataset.
Overall, baselines that capture the conversational discourse structures show better performances, such as \emph{PtrNet}, \emph{DiaBERT}, and \emph{Struct}.
\emph{Struct}, in particular, shows impressive performance by specifically modeling the speaker-utterance and speaker-mentioning structure features, yielding superior outcomes on most metrics.
However, despite the strengths of the existing models, our proposed framework surpasses them all under cluster-level evaluation metrics.
We've achieved enhancements as high as +7.50 on ARI, +3.02 on 1-1, and +4.10 on F1, when compared to \emph{Struct}, the current best-performing model. 
By comparing ours with \emph{Struct} model, we achieve more significant improvement on the exact match evaluation than \emph{Struct}, by a high to 6\% F1 score.
This notable progression is mainly attributed to our distinctive approach of incorporating utterance-distance and partial-replying structures into the model.
Our framework's ability to capture and utilize such dynamic contextual discourse features further underlines their importance in dialogue disentanglement.
In Table \ref{table:addexp1}, we further give the performances under the pairwise link-matching perspective.
We see that our system still outperforms the best-performing baseline.
The above comparisons demonstrate the efficacy of our overall system.

Moving to a different corpus, we present our model's performance on the Movie Dialogue dataset in Table \ref{table:addexp2}. 
Here, given that each session in this dataset only features a single thread, we solely report the cluster-level results.
As seen, our model boosts the baselines on this dataset much more strikingly, with an increase of 8.60(=66.10-57.50) NMI,
20.12(=58.32-38.20) ARI and 8.36(=83.06-74.70) S-F, respectively.
This substantial improvement is partly due to the Movie Dialogue dataset's unique structure, which has fewer utterances per dialogue compared to the Ubuntu IRC data.
The limited conversational contexts available for each utterance pose a unique challenge that our model adeptly handles.
This capability again highlights the versatility and adaptability of our framework in handling different dialogue structures and datasets.
Even when compared to the robust \emph{Struct} model, our framework, equipped with rich structural features, first-order decoding, and bidirectional contextual encoding, consistently delivers superior results.
This comparative analysis unequivocally demonstrates our system's efficacy and potential for further applications in dialogue disentanglement.

\begin{table}[!t]
\fontsize{9.5}{11.5}\selectfont
\setlength{\tabcolsep}{6.mm}
    \caption{
    Pair-wise link matching on Ubuntu IRC data.
    }
    \centering
    \begin{tabular}{llll}
    \hline
   \textbf{ Methods }  & \textbf{ P } & \textbf{ R } & \textbf{ F1 } \\
    \hline
   Transition~\cite{liuSGLWZ20-ijcai} & 62.16 & 20.58 & 49.70   \\
    BERT+MF~\cite{zhu-etal-2021-findings} & 73.90 & 71.30 & 72.60 \\
    PtrNet \cite{yu-joty-2020-online} & 74.70 & 72.70 & 73.68 \\
             \quad +Joint train  & 74.00 & 71.30 & 72.70 \\
             \quad +Self-link  & 74.80 & 72.70 & 73.73 \\
             \quad +Joint train & 74.50 & 71.70 & 73.10 \\
    \cdashline{1-4}
    \textbf{Ours} & \textbf{76.58} & \textbf{73.72} & \textbf{75.07} \\
    \hline
    \end{tabular}
\label{table:addexp1}
\end{table}

\begin{table}[!t]
\fontsize{9.5}{11.5}\selectfont
\setlength{\tabcolsep}{6.mm}
    \caption{
    Cluster-level results on Movie Dialogue data.
    }
    \centering
    \begin{tabular}{lccc}
    \hline
   \textbf{Methods}  & \textbf{NMI} & \textbf{ARI} & \textbf{S-F} \\
    \hline
     Transition~\cite{liuSGLWZ20-ijcai} & 32.90 & 24.80 & 65.00    \\
     DiaBERT~\cite{tiandali-arxiv} & 36.20 & 18.00 & 49.70    \\
     Struct~\cite{aclMa0Z22} & 44.80 & 32.70 & 69.50   \\
     Bi-Level~\cite{chengyuhuang-emnlp} & 57.50 & 38.20 & 74.70   \\
    \cdashline{1-4}
     \bf Ours &  \textbf{66.10} & \textbf{58.32} & \textbf{83.06}  \\
    \hline
    \end{tabular}
\label{table:addexp2}
\end{table}

\subsection{Model Ablation}

In our overall framework, the rich discourse features and optimization strategies are carefully designed to synergize with each other, with the goal of enhancing the model's capacity for dialogue disentanglement.
To investigate and quantify the individual contribution of each component, we have conducted comprehensive ablation studies.
Table \ref{table:ablation} presents a detailed summary of these analyses.

The four types of discourse structures incorporated into our framework — namely the utterance-distance structure ($G^D$), partial-replying structure ($G^R$), speaker-utterance structure ($G^S$), and speaker-mentioning structure ($G^M$) — contribute significantly to the overall performance of the model.
In particular, the removal of any one of these structures invariably leads to a decline in performance, indicating their non-trivial roles in the task.
However, it's worth noting that the impacts of these structures vary.

The dynamic structures, namely the utterance-distance and partial-replying structures, exert more influence on the model's performance than their static counterparts, the speaker-utterance, and speaker-mentioning structures.
This distinction is more pronounced in the fine-grained pair-wise relation detection task, where the speaker-mentioning and partial-replying structures play especially pivotal roles.
Of all the four structures, the partial-replying structure has the most pronounced impact on the performance.
The removal of all structure information from the model precipitates the most severe decline in performance, underscoring the overarching importance of these discourse structures in dialogue disentanglement.

Further, we inspect the influence of the optimization strategies.
The hierarchical ranking loss, which involves the integration of $\mathcal{L}_2$ and $\mathcal{L}_3$, is instrumental in enhancing the model's performance.
The exclusion of either $\mathcal{L}_3$ or both $\mathcal{L}_2$ and $\mathcal{L}_3$ from the training phase leads to a noticeable dip in the overall cluster-level evaluation results.
This indicates that the hierarchical optimization of the task is a crucial factor in achieving high performance.

Finally, the significance of the decoding method becomes evident when we switch from our proposed easy-first decoding mechanism to the standard front-to-back approach.
A consistent decrease in performance across both the cluster-level and pair-wise metrics confirms the necessity of a more advanced decoding method for this task.
This finding reinforces the notion that the decoding stage in dialogue disentanglement should not be overlooked, and that an efficient decoding strategy can substantially contribute to the overall success of the model.

\begin{table}[!t]
\fontsize{9.5}{11.5}\selectfont
\setlength{\tabcolsep}{5.7mm}
\caption{
Ablation results (cluster-level metrics).
`Seq. order' means using the sequential front-to-back decoding order.
}
\centering
\begin{tabular}{l cccccc}
\hline
  \multicolumn{1}{c}{\multirow{2}{*}{\textbf{Methods}}}  & \multicolumn{4}{c}{ \textbf{Cluster-level}} & \multicolumn{1}{c}{ \textbf{Pair-wise} } \\
        \cmidrule(r){2-5}\cmidrule(r){6-6}
 & \textbf{VI} & \textbf{ARI} & \textbf{1-1} & \textbf{F1} & \textbf{F1} \\
\hline
\textbf{Ours} (Full) & \textbf{94.23} & \textbf{81.10} & \textbf{83.90} & \textbf{47.63}  & \textbf{75.07} \\
\hdashline
\multicolumn{5}{l}{\textbf{$\bullet$ Discourse structure}}\\
\quad w/o $G^S$ & 93.98 & 79.20 & 82.78 & 46.51  & 74.58 \\
\quad w/o $G^M$ & 93.88 & 78.76 & 83.08 & 46.24  & 73.71 \\
\quad w/o $G^D$ & 94.12 & 78.00 & 80.08 & 45.12 & 74.69 \\
\quad w/o $G^R$ & 94.00 & 76.83 & 82.92 & 44.68  & 73.39 \\
\quad w/o ALL & 93.45 & 76.20 & 80.43 & 45.23  & 72.77 \\
\hdashline
\multicolumn{5}{l}{\textbf{$\bullet$ \emph{Hierarchical ranking loss}}}\\
\quad w/o $\mathcal{L}_3$ & 94.04 & 80.74  & 83.46 & 46.92  & 74.53 \\
\quad w/o $\mathcal{L}_2$\&$\mathcal{L}_3$ & 93.76 & 79.63 & 82.08 & 45.43  & 74.70 \\
\hdashline
\multicolumn{5}{l}{\textbf{$\bullet$ \emph{Easy-first decoding}}}\\
\quad $\to$ Seq. order & 93.67 & 79.38 & 82.34 & 45.32  & 74.11\\
\hline
\end{tabular}
\label{table:ablation}
\end{table}

\subsection{In-depth Analyses}

Above we have demonstrated the efficacy of our overall framework.
Now we take a further step, presenting more in-depth analyses to answer the following several questions and explore how our method advances the task.



\paratitle{Q1: What is the influence of utterance distance on the dialogue disentanglement task?}

\begin{figure}[!t]
\includegraphics[width=0.65\columnwidth]{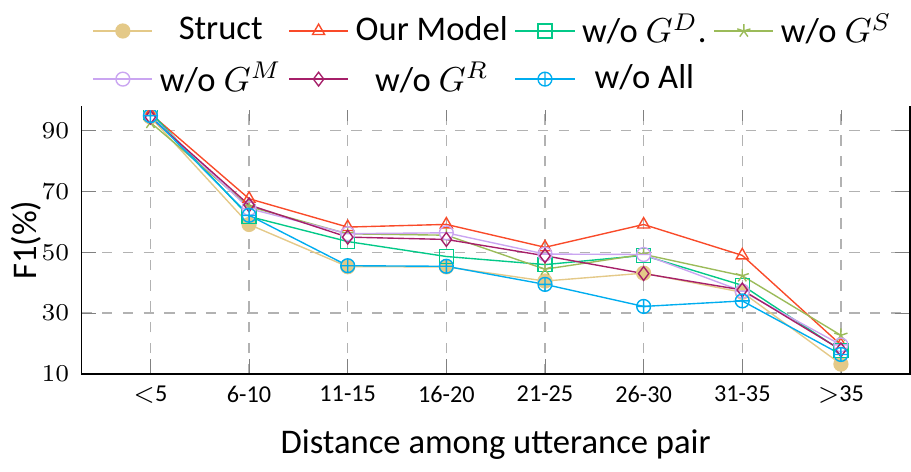}
\caption{Performance under varying utterance distances.
}
\label{fig:distance_exp}
\end{figure}

In the field of discourse analysis, the distance between utterances is a critical factor that affects the performance of disentanglement models.
The complexity of the task is inherently impacted by the utterance distance, with greater distances posing more challenges in identifying the relationships between utterances.
To quantitatively assess the impact of varying utterance distances on dialogue disentanglement, we conducted a rigorous comparison between our method, the \emph{Struct} baseline, and different variants of our model that exclude specific types of structural features, as shown in Figure~\ref{fig:distance_exp}.

The results of our experiments provided valuable insights.
As anticipated, we consistently observed a decrease in performance across all methods as the utterance distance increased.
This finding not only confirms our initial hypothesis that larger utterance distances present a more challenging learning context but also underscores the importance of incorporating appropriate structural features to address this problem.

Both the \emph{Struct} baseline and our models exhibited resilience against the performance degradation caused by increased utterance distances, particularly when compared to the variant model that lacked any discourse structure features (w/o All).
This resilience clearly demonstrates the effectiveness of discourse structure features in mitigating the challenges posed by long-range dependencies \cite{chen-2022-modeling}.
Furthermore, our comprehensive model outperformed the Struct baseline, primarily due to the integration of diverse and advanced structural discourse features.

Upon closer examination of the individual features in our models, we discovered that the Gaussian-based utterance-distance structure feature had a significant impact on shorter distances ($\le$25), while the partial-replying structure feature played a crucial role in addressing longer distances ($>$30).
This fine-grained analysis provides valuable insights into the unique contributions of each discourse feature, highlighting their respective strengths and applications in dialogue disentanglement tasks.

\paratitle{Q2: How does the session size influence the disentanglement task?}

\begin{figure}[!t]
\includegraphics[width=0.75\columnwidth]{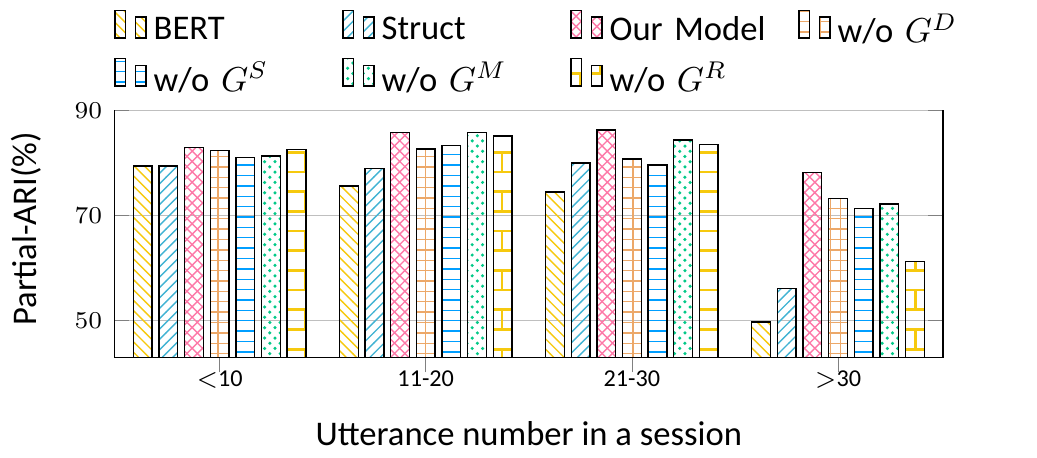}
\caption{Results with varying session sizes.
}
\label{fig:session_exp}
\end{figure}

Increasing the session numbers within a conversation invariably adds to the complexity of topic entanglement, which subsequently makes the disentanglement task more challenging.
To evaluate the impact of different session sizes, we utilized a novel metric, partial-ARI, which is defined in Eq.\eqref{eq:pari}.
As shown in Figure~\ref{fig:session_exp}, this metric enables the assessment of the matching effect of sessions with varying sizes.

The experimental results reveal the influences of session size on the dialogue disentanglement task.
When the dialogue comprised less than 30 sessions, the effect of session numbers appeared to be somewhat minimal.
However, the BERT model, which operates without any discourse structure information, underperformed compared to the \emph{Struct} model equipped with two static conversational structure features.

When the session size increased beyond 30, there was a noticeable decrease in performance across all models.
Despite this overall decline, our models still provided two noteworthy observations.
Firstly, our overall framework demonstrated impressive capability when dealing with complex dialogues, surpassing other baselines.
Secondly, the partial-replying structure feature exhibited a vital role in counteracting the negative impacts of increased session size.

By providing direct signals of intrinsic relations of utterances, the partial-replying structure feature significantly enriches the semantic features available for the task, thereby maintaining the model's performance even under challenging conditions.
This finding underscores the importance of this feature in developing robust dialogue disentanglement models, particularly when faced with conversations characterized by large session sizes.

\paratitle{Q3: How does the Hierarchical Ranking Loss (HRL) mechanism enhance dialogue disentanglement?}

The Hierarchical Ranking Loss (HRL) mechanism, introduced in our method, is designed specifically to optimize cluster-level predictions.
The quantitative results in previous sections have already demonstrated the effectiveness of HRL.
Now, we dive deeper into an in-depth exploration of how the HRL mechanism aids in improving performance.

\begin{figure}[!t]
\centering
\subfloat[Prediction changing ratio]{{\includegraphics[width=0.38\columnwidth]{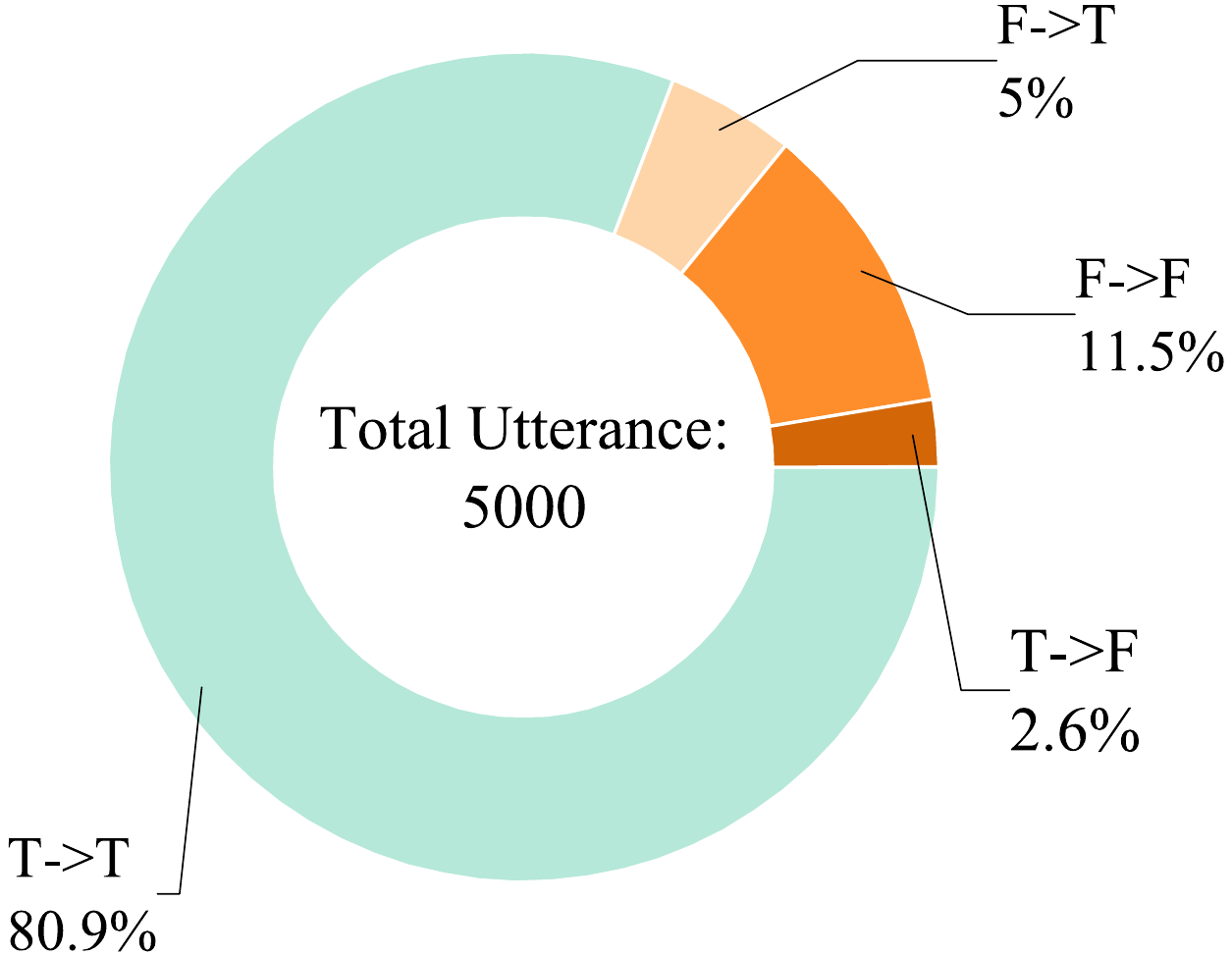} }}%
\subfloat[Change of utterance levels]{{\includegraphics[width=0.38\columnwidth]{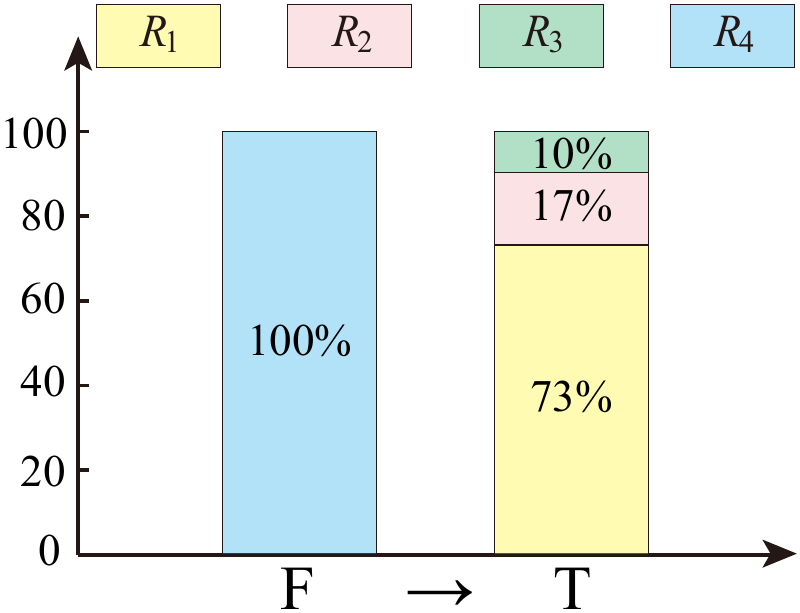} }}%
\caption{
Decomposing the predictions.
X$\to$Y represents the change of prediction without and with equipping with the HRL mechanism.
For example, F$\to$T means that the false (F) prediction on an instance by our model without HRL turns correct (T) when using HRL.
Similarly for F$\to$F, T$\to$T, T$\to$F.
}%
\label{fig:setwise-exp}%
\end{figure}

To elucidate the contribution of HRL, we analyze the changes in the cluster-level true and false predictions of our system when equipped with the HRL method compared to the system devoid of HRL.
The changing ratio of predictions depicted in Figure~\ref{fig:setwise-exp}(a) indicates that the HRL method has corrected approximately 5\% of predictions out of a total of 5k instances.
This improvement by itself is a substantial indication of the effectiveness of the HRL method in enhancing the model's overall performance.

To understand the specifics of this improvement, we look closely at the positive changes, i.e., false predictions corrected to true (F$\to$T).
We categorize these corrected predictions based on the detailed utterance types to comprehend the underlying mechanism of performance improvement.
According to the data shown in Figure~\ref{fig:setwise-exp}(b), it is clear that a significant proportion (73\%) of $R_4$ utterances have been corrected to the exact parent utterances.
This correction remarkably enhances both cluster-level and pair-wise prediction accuracies.
Moreover, an additional 17\% and 10\% of the $R_4$ utterances have been shifted to the ancestor $R_2$ utterances and the inner-session $R_2$ utterances, respectively.
This change, while more subtle, significantly contributes to the improvement of cluster-level prediction.
In summary, the HRL mechanism demonstrates a significant capacity to rectify prediction errors, thereby enhancing the model's overall performance.
The detailed analysis of the types of utterances that are most affected by the application of HRL offers valuable insights into its inner workings and effectiveness, further substantiating its value in dialogue disentanglement tasks.


\begin{figure}[!t]
\includegraphics[width=0.65\columnwidth]{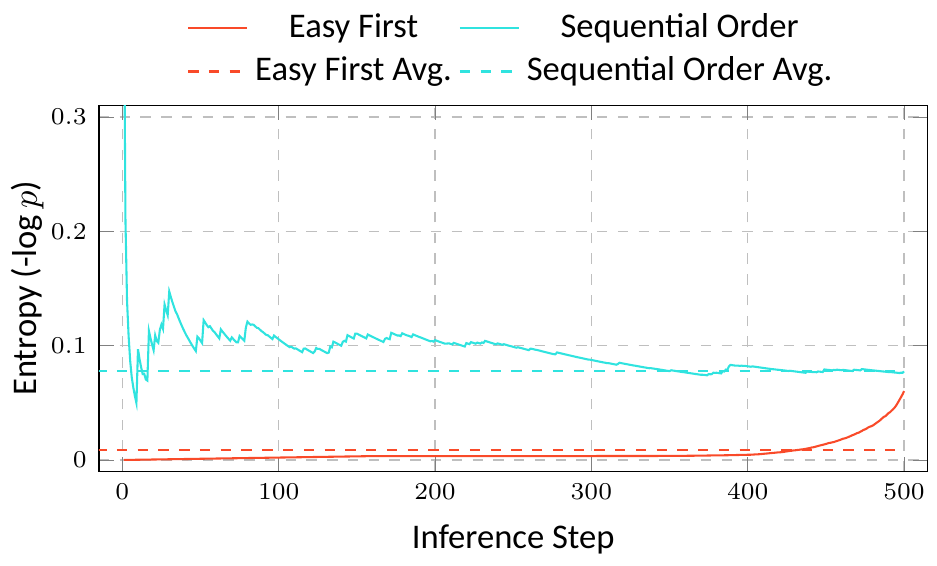}
  \caption{
The trajectory of model prediction entropy on decoding the dialogue utterances with different decoding strategies.
Lower entropy values indicate higher predicting certainty, i.e., easier decision.
  }
  \label{fig:easy_first_res}
\end{figure}

\paratitle{Q4: What is the working mechanism of the easy-first decoding strategy in enhancing the performance of dialogue disentanglement?}

In this part of the analysis, our objective is to uncover the underlying mechanism that allows the easy-first decoding strategy to bolster overall performance in the task of dialogue disentanglement. 
To this end, we examine the decoding behavior of our model by measuring the entropy of the model prediction at each utterance decoding step. 
Intuitively, a lower entropy ($\downarrow-\log p$) corresponds to higher predicting certainty ($\uparrow p$), which suggests that the decoding decision is easier to make with confidence.

We make a comparative study against the standard sequential decoding strategy, which operates in front-to-back order. 
As shown by the trends plotted in Figure~\ref{fig:easy_first_res}, the proposed easy-first decoding strategy demonstrates significant superiority in coordinating task inference. 
To elaborate, in the initial 400 steps, the easy-first decoding strategy attains a near-perfect predicting certainty ($p\sim$1). 
This implies a precise and highly confident decision-making process in determining the replying relation between utterances.
In contrast, the sequential-order decoding strategy showcases a rather fluctuating process, particularly in the initial steps. 
The reason behind this can be attributed to the sequential-order paradigm, wherein available features from only prior contexts can be severely limited for reasoning the correct parent for the current utterance. 
This constraint considerably hampers the ability of the model to make accurate and confident decisions.
In essence, the easy-first decoding strategy, by ensuring high predicting certainty from the initial steps, provides a decisive edge in the task of dialogue disentanglement. 
Its performance, as compared to the sequential decoding strategy, validates its effectiveness in providing clear and confident decisions for identifying replying relations between utterances.

\subsection{Q5: How do LLMs compare with fine-tuned models in dialogue disentanglement?}

In recent years, there has been a surge in the development and utilization of generative-based large language models (LLMs), which have yielded promising results in various Natural Language Processing (NLP) tasks.
This substantial advancement in LLMs, backed by innovations such as in-context learning~\cite{brolma-nips-2020}, prompt-based learning~\cite{liuppa-acm-2023}, and the chain-of-thought approach~\cite{weicpe-nips-2022}, has further enhanced the adaptability of these models to downstream tasks.
In the wake of the introduction of ChatGPT, the GPT-3.5 model, along with its derivative versions, has, remarkably, demonstrated performance equivalent to or surpassing that of fine-tuned models in a variety of NLP tasks, even in zero-shot settings.
These tasks encompass areas such as sentiment analysis, question answering, and summarization~\cite{qinisa-arr-2023}.

\begin{figure}[!t]
\includegraphics[width=0.99\columnwidth]{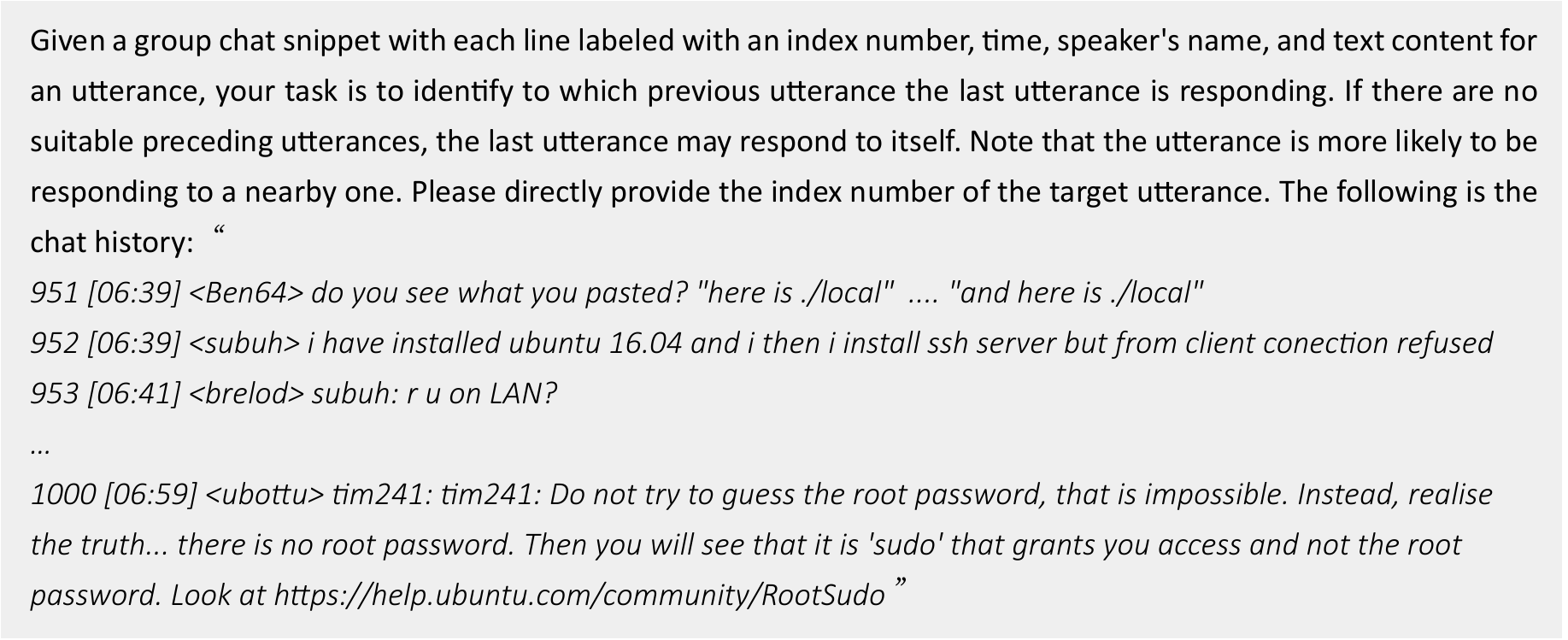}
  \caption{
  A prompt to request ChatGPT for dialogue disentanglement.
  }
  \label{fig:prompt}
\end{figure}

This leads us to question the necessity and relevance of employing fine-tuned models in the specific task of dialogue disentanglement in the LLM era.
Thus far, there is lacking a research exploration regarding the application of LLMs in dialogue disentanglement tasks. 
Motivated by this, we ventured to design an exploratory experiment aimed at comparing the performance of fine-tuned models with the GPT-3.5 model in the context of dialogue disentanglement.
Due to the unavailability of access to fine-tune the GPT-3.5 model, we employed the GPT-3.5-Turbo-0301 API~\footnote{\url{ https://platform.openai.com/docs/models/gpt-3-5}} to execute zero-shot experiments.
We carefully selected 100 utterances randomly from the test set, which encompassed the 50 sentences preceding each utterance.
The prompt was employed to instruct the GPT-3.5 model to identify a 'parent' utterance, as depicted in Figure~\ref{fig:prompt}.



\begin{table}[!t]
\fontsize{9.5}{11.5}\selectfont
\caption{
Comparison of our model and GPT-3.5 API.
}
\centering
\begin{tabular}{cc}
\hline
Model Name &  Pair-wise Link F1 \\
\hline
Our model &  0.72 \\
GPT-3.5-turbo-03-01 &  0.10\\
\hline
\end{tabular}
\label{table:gpt}
\end{table}

Following this prompt sent to the GPT-API, we received a response indicating the index number of the parent utterance as predicted by ChatGPT.
We then compared the prediction results of the 100 utterances to those of our model.
As shown in Table~\ref{table:gpt}, our fine-tuned model significantly outperformed ChatGPT in a zero-shot setting.
This suggests that LLMs have not yet reached a point where they can match supervised models on all NLP tasks, particularly those involving complex discourse structures.
A possible explanation is that LLMs have not encountered similar tasks during the pre-training stage, leading to subpar performance.
This points to a need for innovative solutions in model design and training strategies to enable LLMs to effectively learn and handle complex discourse features.
In the future, we plan to extend our research to find more sophisticated methods to better integrate LLMs with discourse structures, hoping to achieve superior performance.

\subsection{Case Study}
Finally, to better understand how our overall model helps make correct predictions of dialogue disentangling, we conduct detailed case studies to show the contribution of discourse-aware encoding module and easy-first decoding mechanism, respectively.

\label{Case Study of Dialogue Discourse Structures}
\begin{figure*}[htbp]
\includegraphics[width=0.98\textwidth]{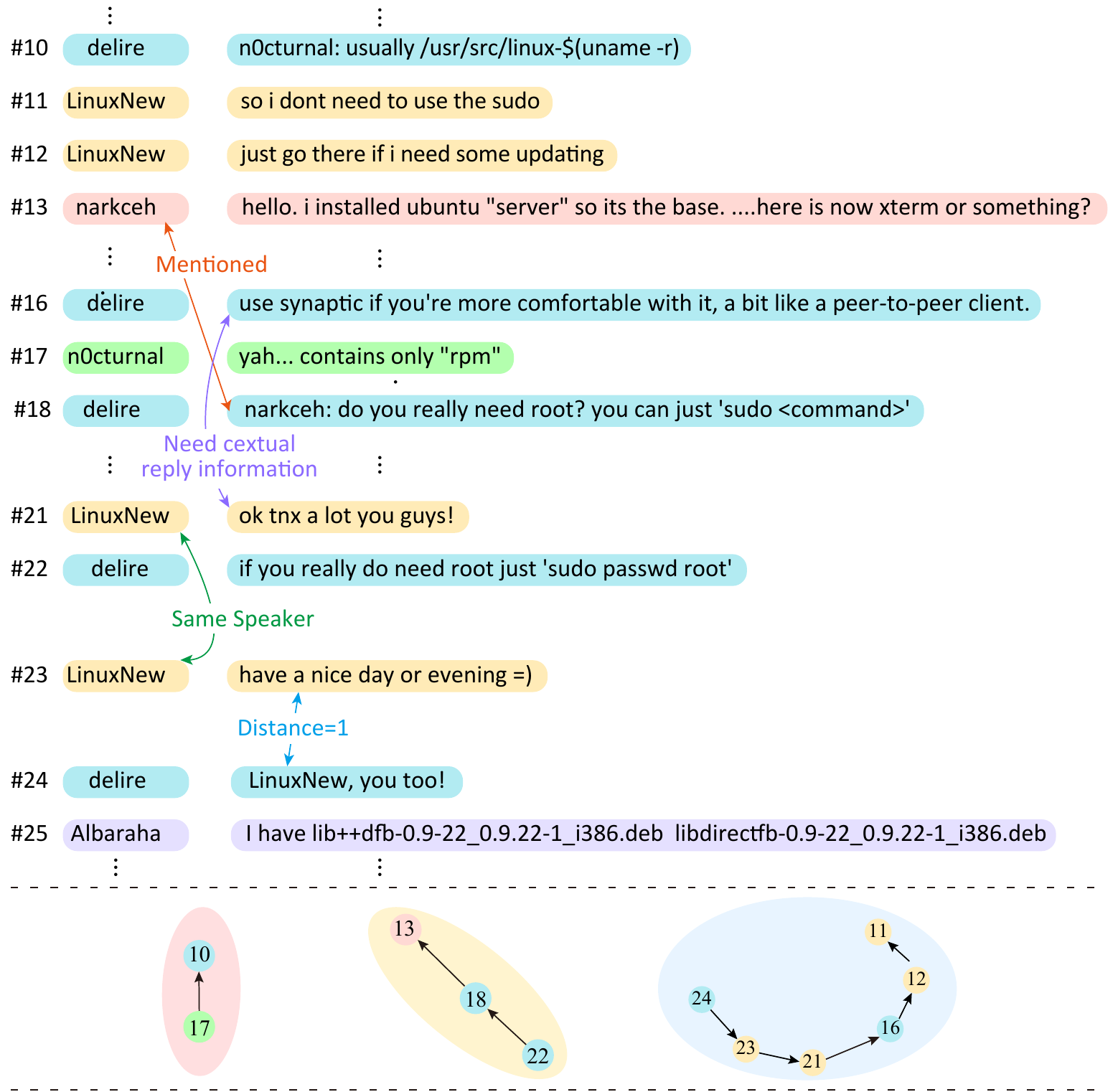}
  \caption{
A case study about how different types of edges help build the reply relationship.
The left of the figure displays a small snippet of a continuous dialogue, with some utterances skipped for clarity.
The cluster of sessions on the right is formed based on the reply relationships between shown utterances.
We also especially highlight how the four different types of edges aid in establishing reply relationships between corresponding utterance pairs.
For instance, the speaker-mentioning edge can provide a non-trivial clue to form the reply relationship between the \#18 utterance and \#13.
}
  \label{fig:case_study_app}
\end{figure*}

\paratitle{Discourse-aware encoding}

As seen in Figure~\ref{fig:case_study_app}, different discourse structural features play varied yet crucial roles in facilitating the replying relation reasoning over the conversation discourse, i.e., by providing rich cross-utterance clues.
For example, the model utilizes the speaker mention information provided by the \#18 utterance to determine that its parent is the \#13 utterance.
In addition to this, the model also determines the \#23 utterance to be a direct reply to the \#22 utterance.
This is primarily due to the fact that the speakers of these utterances are identical, and the contents of the utterances are coherent, making them a likely pair in the reply relation.

Distance information forms another integral part of our model, especially in instances where there are multiple potential candidates. 
For example, the \#24 utterance mentions \textit{LinuxNew}, who is the speaker of the \#21 and \#23 utterances.
If we were to judge solely based on the content of the utterances, the \#24 utterance could potentially be a response to either the \#21 or \#23 utterance.
But our model considers the distance weight, which would naturally be higher for two utterances that are closer in proximity.
This results in a stronger interaction between the \#23 utterance and the \#24 utterance.
Therefore, the model would tend to choose the \#23 utterance as the parent of the \#24 utterance.

Reply information is also a crucial element in our model, especially when it comes to identifying the reply relationship in situations where reference information is insufficient or lacking.
Let's take the \#21 utterance for instance, which is quite unclear to determine its parent utterance in terms of speaker-mention and could generally be a response to several other utterances.
Nevertheless, our model has already established the reply relationship between the \#16 and \#12 utterances.
Coupled with the fact that the \#21 utterance shares the same speaker as the \#12 utterance, the model concludes that it is highly likely for the 21st utterance to be a reply to the \#16 utterance.
The ability to accurately infer such reply relationships is vital in real-life conversation scenarios characterized by brevity and precise exchanges.
These are precisely the conditions under which our model operates and excels, effectively recognizing and mapping dynamic reply relationships.
By understanding and utilizing this dynamic, our model successfully gathers more precise contextual information, leading to more effective dialogue disentanglement.

\begin{figure*}[!t]
\includegraphics[width=0.98\textwidth]{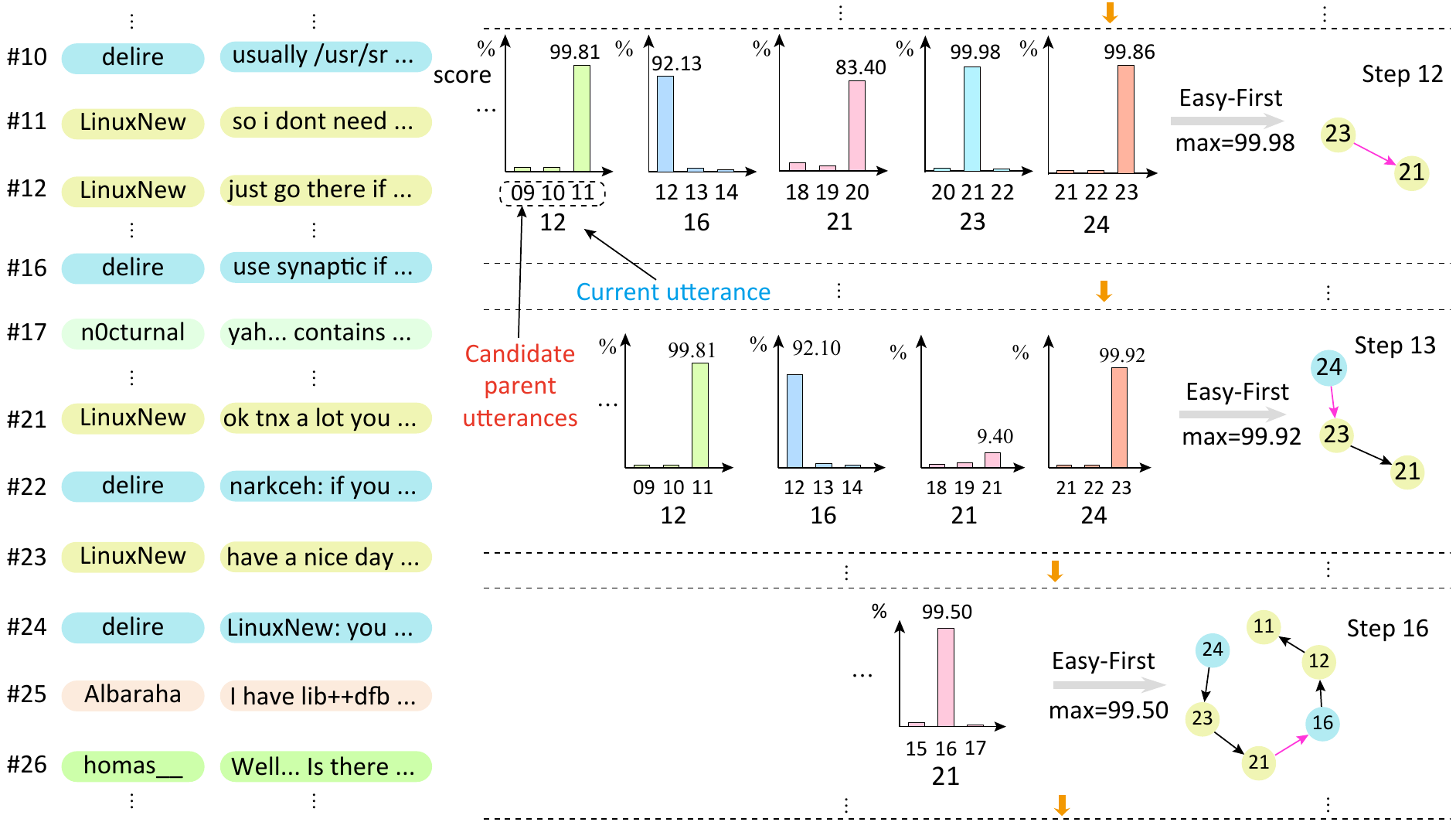}
  \caption{
Qualitative analysis of how our system predicts the reply-to structures (predictions are correct).
We show the easy-first decoding process (right) based on the input dialog (partially shown on the left) that is randomly picked from the Ubuntu IR testing set.
`\textcolor{ipink}{$\rightarrow$}' denotes the decision of current replying relation.
Appendix $\S$\ref{Case Study of Dialogue Discourse Structures} visualizes the conversational discourse structures.
  }
  \label{fig:case_study}
\end{figure*}

\paratitle{Easy-First decoding}
In this section, we employ a case study to demonstrate the workings of the easy-first decoding approach, specifically focusing on representative steps in a testing dialogue instance where our system flawlessly predicts all replying relations.
As seen in Figure~\ref{fig:case_study}, our system cleverly makes each decoding decision by selecting the current most confident one.
For instance, at the significant juncture of step 13, the pairs that should be the simplest to decode would ideally be \#24$^{\curvearrowright}$\#23, primarily because \#24 references \textit{LinuxNew}, the speaker name of \#23.
Leveraging such clues, our system correctly predicts the replying relation at this stage.
With each passing iteration, the model outputs the reply pair with the highest score, actively incorporating it into the partial replying graph, which is used in the subsequent round of computation.

Basing its predictions on the graph it has thus far constructed, our model consistently provides increasingly confident and precise predictions, even in the face of more complex cases.
For instance, during the 16-th step, the model capitalizes on the dynamic replying context to accurately predict the \#16 utterance as the parent of the \#21 utterance.

In conclusion, our system demonstrates its proficiency in dealing with varying complexities of decoding and predicting replying relations in a dialogue.
Through the utilization of strategic structural features and dynamic replying context, it navigates the discourse with increasing confidence and precision, making it a promising approach in the field of dialogue disentanglement.


\section{Discussions}
\subsection{Conclusion}
In this study, we rethink the discourse attribute of conversations, and improve the dialogue disentanglement task by taking full advantage of the dialogue discourse.
From the feature modeling perspective, we build four types of dialogue-level discourse graphs, including the static speaker-role structures (i.e., speaker-utterance and speaker-mentioning structure), and the dynamic contextual structures (i.e., the utterance-distance and partial-replying structure).
We then develop a structure-aware framework to encode and integrate these heterogeneous graphs, where the edge-aware graph convolutional networks are used to learn the rich structural features for better modeling of the conversational contexts.
From a system optimization perspective, we first devise a hierarchical ranking loss mechanism, which groups the candidate parents of current utterance into different discourse levels, and carries model learning under three hierarchical levels, for both pair-wise and session-wise optimizations.
Besides, we present an easy-first decoding algorithm, which performs utterance pairing in an easy-to-hard manner with both the precedent and subsequent context.
Experimental results on two benchmark datasets show that our overall system outperforms the current best results on all levels of evaluations.
Via further analyses, we demonstrate the efficacy of all the above-proposed designs, and also reveal the working mechanism of how they help advance the task.

\subsection{Future Work}

Although our model has achieved competitive performance in the dialogue disentanglement task, there is still room for improvement.
We believe the model can be improved in the following directions:

\textbf{1) Exploitation of Fine-grained Discourse Features}:
In this study, we have extensively considered utterance-level dialogue features and achieved promising results. 
However, we have not yet fully exploited fine-grained features, which has limited the further optimization of the model.
In fact, fine-grained features, such as syntactic structure and coreference resolution~\cite{haofei-global}, are very beneficial to the dialogue disentanglement task.
In the future, if we can integrate sentence-level and fine-grained discourse features for achieving global-local feature mining, the effectiveness of dialogue disentanglement will be further enhanced.

\textbf{2) Integration with Dialogue Generation Tasks}:
The main focus of this study is the dialogue disentanglement task, with the goal of processing existing dialogues.
In contrast, the dialogue generation task aims to generate new dialogues based on existing ones.
These two processes are reversible and interdependent.
Improved dialogue disentanglement can facilitate generation~\cite{ zhesdr-aaai-2021}, and conversely, superior generation performance indicates a deeper understanding of dialogue disentanglement.
Therefore, integrating these two tasks would likely promote the synchronous enhancement of both tasks' performance.

\textbf{3) Incorporation of Large Language Models}:
Large language models have demonstrated superiority in various natural language processing tasks.
However, the experiments in this study have found that their performance in the dialogue disentanglement task remains subpar, indicating that the potential of large language models has not yet been fully exploited.
In the future, we might consider using in-context learning~\cite{ brolma-nips-2020} techniques to enable the model to learn dialogue disentanglement tasks, or apply LoRA~\cite{ hulla-iclr-2022} technology to fine-tune the model for this task, thereby fully exploiting the potential of large language models in dialogue disentanglement tasks.
This approach would not only enhance the effectiveness of dialogue disentanglement tasks but also broaden the application scenarios of large language models, contributing to the realization of a universal natural language processing model.

\bibliographystyle{ACM-Reference-Format}
\bibliography{acmbib}
\appendix

\end{document}